\pdfoutput=1
\documentclass[10pt,logo,copyright]{techreport}
\usepackage[authoryear,sort&compress,round]{natbib}

\usepackage[utf8]{inputenc}
\usepackage[T1]{fontenc}
\usepackage{fancyhdr}
\usepackage{ifthen}
\usepackage{times}
\usepackage{latexsym}
\usepackage{xspace}

\usepackage{pifont} 
\usepackage[most]{tcolorbox}
\usepackage{enumitem}
\usepackage[table]{xcolor}
\usepackage{subcaption}
\usepackage{amssymb}
\usepackage{array}
\usepackage{longtable}

\usepackage{listings}
\usepackage{xcolor}
\usepackage{textcomp}

\usepackage{parskip}
\usepackage{url}
\usepackage{hyperref}
\usepackage{booktabs}
\usepackage{amsfonts}
\usepackage{nicefrac}
\usepackage{microtype}
\usepackage[dvipsnames]{xcolor}
\usepackage{graphicx}
\usepackage{tabularx}
\usepackage{makecell}
\usepackage{float}
\usepackage[section]{placeins}
\usepackage{wrapfig}

\usepackage{amsmath,amsfonts,bm,bbm}
\usepackage{multirow}

\definecolor{best_green}{HTML}{DDF4DD}
\definecolor{second_purple}{HTML}{E4E0FF}
\definecolor{figcitecolor}{HTML}{30b776}

\usepackage[T1]{fontenc}
\usepackage[utf8]{inputenc}
\usepackage{microtype}
\usepackage{inconsolata}

\usepackage{amsmath}
\usepackage{amssymb}
\usepackage{booktabs}
\usepackage{graphicx}
\usepackage{xspace}
\usepackage{titletoc}

\hypersetup{
    colorlinks=true,
    linkcolor=figcitecolor,
    citecolor=tongyipurple,
    urlcolor=tongyicite
}

\definecolor{best_green}{HTML}{DDF4DD}
\definecolor{second_purple}{HTML}{E4E0FF}
\definecolor{deltapurple}{RGB}{81,45,168}
\definecolor{deltagreen}{RGB}{0,100,60}
\definecolor{refgray}{gray}{0.45}
\definecolor{findingbg}{HTML}{E8E9F2}
\definecolor{findingpurple}{HTML}{584489}

\newcommand{\best}[1]{\cellcolor{best_green}\textbf{#1}}
\newcommand{\second}[1]{\cellcolor{second_purple}\underline{#1}}
\newcommand{\posdelta}[1]{\textcolor{deltapurple}{#1}}
\newcommand{\negdelta}[1]{\textcolor{deltagreen}{#1}}
\newcommand{\refcell}[1]{\textcolor{refgray}{#1}}

\newcommand{\besttext}[1]{%
  {\setlength{\fboxsep}{0.6pt}\colorbox{best_green}{#1}}%
}
\newcommand{\secondtext}[1]{%
  {\setlength{\fboxsep}{0.6pt}\colorbox{second_purple}{#1}}%
}

\newcommand{\modelname}{\texttt{\textcolor{tongyipurple}{\textbf{InnerZoom}}}\xspace}

\newtcolorbox{findingbox}{
  colback=findingbg,
  colframe=findingpurple,
  boxrule=0.5pt,
  arc=1.5pt,
  left=6pt,
  right=6pt,
  top=5pt,
  bottom=5pt,
  before skip=6pt,
  after skip=8pt
}

\title{One Forward Beats Two: InnerZoom for Accurate and Efficient GUI Grounding}
\vspace{-1.0em}
\author{
Chen Liu$^{2*}$,
Ling Chen$^{2}$,
Hanzhang Zhou$^{1\dagger}$,
Liangyu Chen$^{1}$,
Chenglin Cai$^{1}$,
Xin Yu$^{3}$,
Steven Hoi$^{1}$,
Yue Wang$^{1\dagger}$
}
 
\begin{document}
\maketitle
\begin{figure*}[htp]
    \centering
    \includegraphics[width=1.0\linewidth]{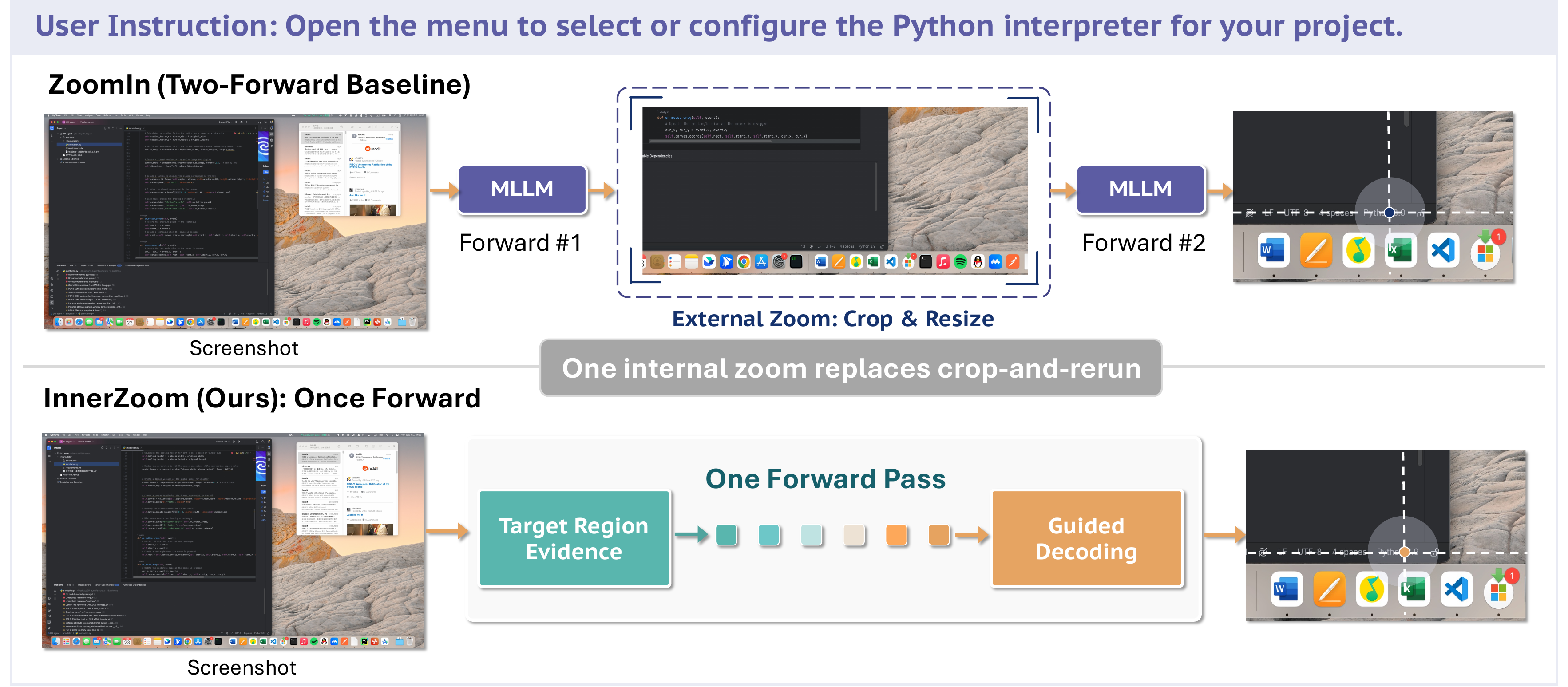}
    \caption{
    \textbf{InnerZoom performs external zoom with one forward pass.}
    ZoomIn-style methods predict a coarse target region, externally crop and resize it, and perform a second forward pass to generate coordinates, increasing latency and computational cost. \textcolor{tongyipurple}{\textbf{InnerZoom}} instead reuses internal target-region evidence to guide coordinate decoding within a single forward pass. Under the same 4B training configuration, \textcolor{tongyipurple}{\textbf{InnerZoom}} surpasses ZoomIn on UI-Vision, OSWorld-G-Refine, and OSWorld-G while using only one forward pass.
    }
    \vspace{-1.0em}
    \label{fig:teaser}
\end{figure*}

\begin{abstract}

MLLM-based GUI grounding methods commonly formulate target localization as autoregressive coordinate generation, enabling models to leverage the strong instruction-following and semantic understanding capabilities of MLLMs. 
However, this formulation requires the model to retain region-level target evidence while decoding coordinate tokens with the spatial precision demanded by GUI clicking. 
Our diagnostic analysis reveals that target-region awareness emerges in intermediate decoder layers but is neither retained nor translated into the final coordinate prediction.
Existing ZoomIn-style methods address this issue through an external crop-and-rerun pass, which improves localization but increases end-to-end latency and computational cost. 
To retain the accuracy benefits of two-pass zooming without this extra cost, we propose \modelname, a single-forward framework for cross-layer evidence bridging. 
\modelname transforms target-related cues from the original forward pass into a compact cross-layer evidence state, then preserves, refines, and reinjects this state throughout later decoding layers to guide coordinate prediction.
Extensive experimental results suggest that \modelname-4B achieves state-of-the-art performance on all six GUI grounding benchmarks, obtaining 64.7 on OSWorld-G, 40.2 on UI-Vision, 73.1 on OSWorld-GR, and 87.6 on MMBench-GUI, surpassing the previous best results by 4.1, 3.2, 2.9, and 2.3 points, respectively.
Under a controlled 4B setting, \modelname improves the same SFT+RL baseline by 5.3 points on average and outperforms two-pass ZoomIn by 1.3 points on average, while reducing end-to-end latency by up to 31.8\% and TFLOPs by about 29\%. Code and models will be publicly available.

\vspace{-1mm}
\end{abstract}
\abscontent


\section{Introduction}
\label{sec:introduction}

GUI grounding is a fundamental capability for GUI agents, requiring a model to predict an executable click coordinate from a user instruction and a GUI screenshot \citep{nguyen2025gui}.
Recent MLLM-based methods commonly formulate this task as autoregressive coordinate generation, allowing models to leverage the instruction following, semantic understanding, and unified generation capabilities of MLLMs \citep{xu2026mobile,zhou2025mai,li2025screenspotpro}.

However, autoregressive coordinate generation creates a mismatch between region-level visual evidence and point-level coordinate prediction. Precise GUI clicking depends on fine-grained cues within the target region, whereas these cues must be implicitly preserved through decoder states and eventually expressed as discrete coordinate tokens \citep{lin2025boosting,pantazopoulos2025towards}. 
During this conversion, local spatial evidence can fade, causing the model to identify the correct region while still decoding a biased coordinate and producing a near-miss click.
Recent ZoomIn-style methods \citep{zhang2025mvp,jiang2025zoom} alleviate this issue through an external crop-and-rerun procedure. 
As depicted in Figure \ref{fig:teaser}, they first predict a coarse target region, crop and enlarge it from the original screenshot, and then perform a second forward pass to generate the final coordinate. 
Their effectiveness suggests that enhanced target-region evidence benefits point-level grounding. 
However, this external zoom operation requires an additional inference pass and introduces undesirable latency and computation for interactive GUI agents.

\begin{figure*}[htp]
    \centering
    \includegraphics[width=1.0\linewidth]{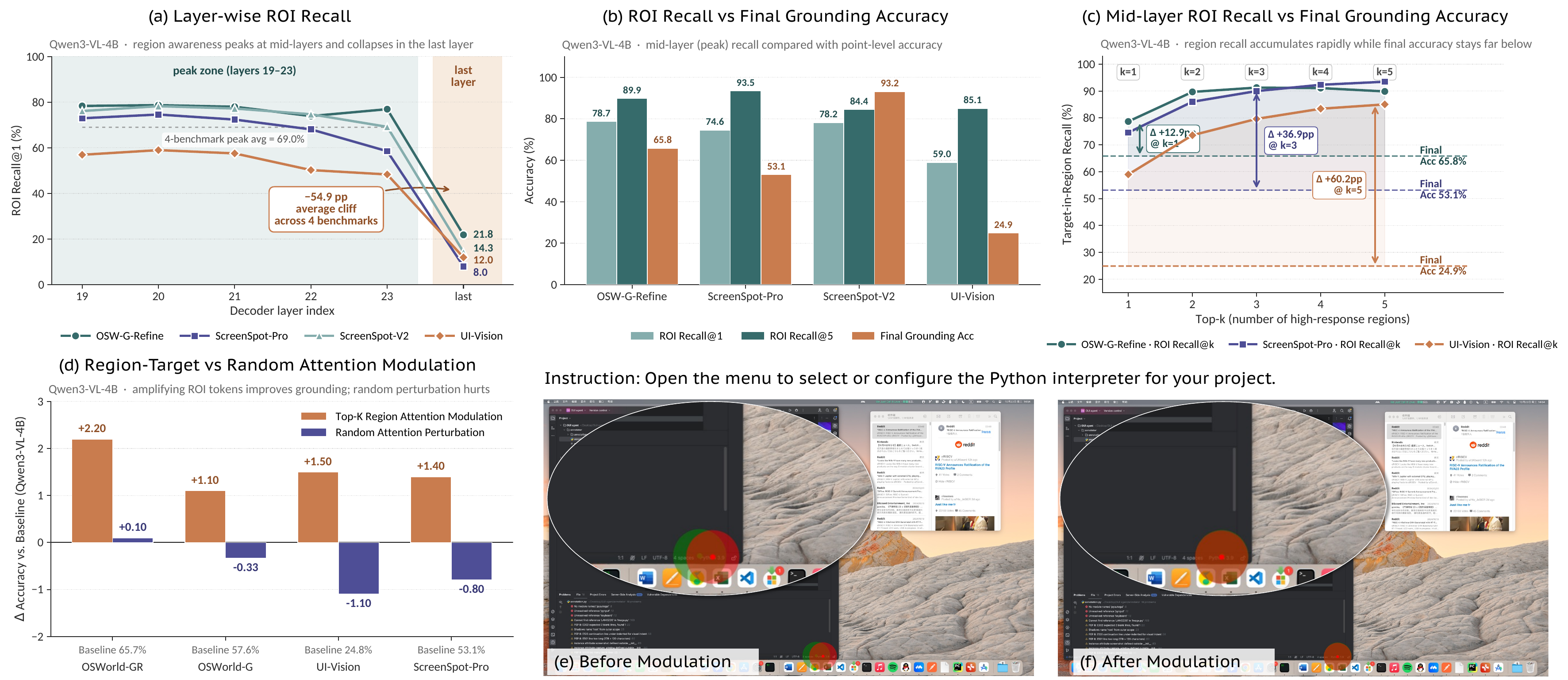}
    \vspace{-1.0em}
    \caption{
    \textbf{Motivation and effect of region-guided evidence modulation.}
    (a) Target-region awareness emerges in intermediate decoder layers but collapses at the final layer, revealing a gap between region-level evidence and point-level decoding.
    (b) High ROI recall does not necessarily translate into high final grounding accuracy across benchmarks.
    (c) Increasing the number of high-response regions rapidly improves target-region recall, while final coordinate accuracy remains much lower.
    (d) Controlled attention modulation shows that amplifying target-region tokens consistently improves grounding accuracy over random-token perturbation.
    (e, f) Region-guided modulation moves the predicted click away from a nearby distractor and toward the intended target. Implementation details are provided in Appendix~\ref{app:attention_intervention}.
    }
    \label{fig:motivation}
\end{figure*}

This raises a natural question: \textit{Does precise GUI grounding really require visual re-observation}?
To answer this question, we analyze text-conditioned visual response maps from intermediate decoder layers. 
As shown in Figure \ref{fig:motivation} (a), localized high-activation regions often emerge around the ground-truth target before final coordinate prediction.
Controlled modulation further suggests that amplifying these regions improves grounding accuracy, whereas an equal-strength random-token intervention provides little benefit or degrades performance, as shown in Figure \ref{fig:motivation} (d)--(f). 
These results indicate that the bottleneck is not simply the lack of visual evidence. 
Instead, intermediate target-region evidence is not preserved and converted into a final point-level click.

To bridge this Region-to-Point Gap, we propose \modelname, a single-forward cross-layer evidence bridging framework for precise GUI grounding. \modelname turns coordinate generation into an evidence-preserving decoding process. 
It identifies target-region evidence from the model’s own text-image responses, converts the corresponding fine-grained visual features into a compact cross-layer evidence state, and progressively refines this state throughout later decoding layers. 
The refined evidence then guides coordinate prediction without external crops, additional test-time inputs, or a second forward pass.
As illustrated in Figure~\ref{fig:teaser}, \modelname performs an internal zoom within one forward pass, replacing the external crop-and-rerun procedure used by ZoomIn-style methods. 
Extensive experiments show that \modelname achieves state-of-the-art performance on five GUI grounding benchmarks. 
Under controlled comparisons with the same 4B backbone and training configuration, it surpasses ZoomIn on UI-Vision \citep{ui_vision}, OSWorld-G-Refine \citep{osworld_g}, and OSWorld-G \citep{osworld_g} while requiring only one forward pass.

Our contributions are as follows.
\begin{itemize}[leftmargin=*,itemsep=0.5pt,topsep=0pt]
    \item We identify the \textbf{Region-to-Point Gap} in MLLM-based GUI grounding. 
    Intermediate decoder layers already contain strong target-region evidence, but this evidence fades during coordinate decoding and is not reliably translated into precise click coordinates.
    \item We propose \modelname, a single-forward cross-layer evidence bridging framework that extracts target-region evidence from intermediate layers, preserves and refines it throughout later decoding, and reinjects it to guide coordinate prediction without external crops or a second forward pass.
    \item Extensive experiments on five GUI grounding benchmarks show that \modelname achieves state-of-the-art performance. 
    Under controlled comparisons with the same 4B backbone and training configuration, it reduces end-to-end latency by 25.5\% to 31.8\% relative to ZoomIn, with a 28.3\% average reduction across four benchmarks, while improving grounding accuracy by up to 4.6 percentage points.

\end{itemize}

\section{Motivation for the Region-to-Point Gap in GUI Grounding}
\label{sec:motivation}

\subsection{Target Evidence Emerges but Does Not Reach the Final Click}
To answer the question raised in the introduction, we analyze text-to-vision response maps from intermediate decoder layers of the Qwen3-VL-4B \citep{Qwen3-VL}. 
Given the response map at decoder layer $l$, we select the top-$k$ high-response visual regions and measure whether they cover the ground-truth click target. 
We refer to this metric as ROI Recall. Details of the extraction procedure are provided in Appendix~\ref{sec:app_more_exp}.

As shown in Figure~\ref{fig:motivation} (a), intermediate decoder layers often identify the target region internally. 
Top-1 ROI Recall peaks at 69.0\% on average across four benchmarks between layers 19 and 23, but drops to only 14.0\% at the final layer, far below the model's earlier target-region awareness. 
Grounding failures therefore do not arise solely from an inability to find the target region.
However, this intermediate evidence fails along two complementary dimensions. 
Across depth, it is not reliably preserved. 
ROI Recall drops by 54.9 percentage points on average from its peak layer to the final layer immediately before coordinate prediction, as shown in Figure \ref{fig:motivation} (a). 
Across the transition from region to point, even the strongest target evidence is not reliably converted into an accurate coordinate. 
At the peak layer, the target is often ranked among the model's high-response regions, yet final grounding accuracy remains much lower. 
On ScreenSpot-Pro, ROI Recall@5 reaches 93.5\% while final accuracy is only 53.1\%. 
On UI-Vision, ROI Recall@5 reaches 85.1\%, but final accuracy is only 24.9\%, as shown in Figure \ref{fig:motivation} (b).

Figure~\ref{fig:motivation} (c) further shows that increasing $k$ rapidly drives ROI Recall toward saturation, while the substantial gap to final coordinate accuracy persists.
This result indicates that the main deficit lies not in covering the target region, but in converting available region-level evidence into a precise point-level action. 
The model thus often identifies the right region in intermediate layers but cannot reliably turn this evidence into a final click. We refer to this discrepancy as the \textbf{Region-to-Point Gap}.

\begin{findingbox}
\textcolor{tongyipurple}{\textbf{Finding 1.}}
Intermediate decoder layers contain strong target-region evidence, but this evidence fades before final coordinate prediction and is not reliably converted into the final click.
\end{findingbox}

\subsection{Steering Attention Toward Target Evidence Recovers Grounding Accuracy}

The analysis above shows that intermediate responses are \emph{correlated} with target regions, but correlation alone does not establish whether they influence the final decision. 
We therefore conduct a controlled attention modulation experiment. 
One intervention amplifies the selected target-region visual tokens, while a matched control amplifies the same number of randomly selected visual tokens with the same modulation strength. 
The two interventions differ only in the visual tokens to which modulation is applied, allowing us to isolate the effect of token selection from modulation magnitude.
Implementation details can be found in Appendix \ref{apd:exp_details}.

As shown in Figure \ref{fig:motivation} (d), amplifying target-region tokens improves grounding accuracy on all four benchmarks, with gains of up to 2.2 percentage points. 
In contrast, amplifying randomly selected visual tokens provides little benefit and can reduce accuracy by up to 1.1 points. 
Since both interventions use the same modulation budget, the gain cannot be attributed merely to increasing attention magnitude. 
Instead, it results from concentrating the modulation on target-region evidence. 
This result suggests that intermediate target-region evidence is decision-relevant and causally influences final coordinate prediction under controlled intervention, rather than being incidentally correlated with the target. 
The qualitative examples in Figure \ref{fig:motivation} (e,f) further illustrate this effect. 
After region-guided modulation, the predicted click shifts toward the intended target.

\begin{findingbox}
\textcolor{tongyipurple}{\textbf{Finding 2.}}
Intermediate target-region evidence is not merely correlated with the target. It causally influences final grounding under controlled intervention and can be actively used to improve coordinate prediction.
\end{findingbox}

Together, Findings 1 and 2 reveal that the bottleneck in GUI grounding is not acquiring target-region evidence. 
The model already forms decision-relevant evidence in intermediate layers, but does not reliably preserve and use it for final coordinate prediction.
Region-guided modulation further confirms that strengthening target-region responses can improve grounding accuracy. However, this inference-time reweighting provides only a transient boost to the selected tokens. 
It does not convert target evidence into an explicit state that persists through subsequent decoder layers. 
As later layers continue to transform the hidden states for coordinate generation, the boosted evidence can again be diluted by competing context and is not explicitly refined for point-level prediction. 
This limitation directly motivates \modelname. 
Rather than applying a transient attention reweighting, \modelname transforms target-region evidence formed within a single forward pass into a compact cross-layer evidence state. 
It preserves, refines, and reinjects this state throughout later decoding layers to provide sustained guidance for coordinate prediction.

\section{Method}
\label{sec:method}

\begin{figure*}
    \centering
    \includegraphics[width=0.98\linewidth]{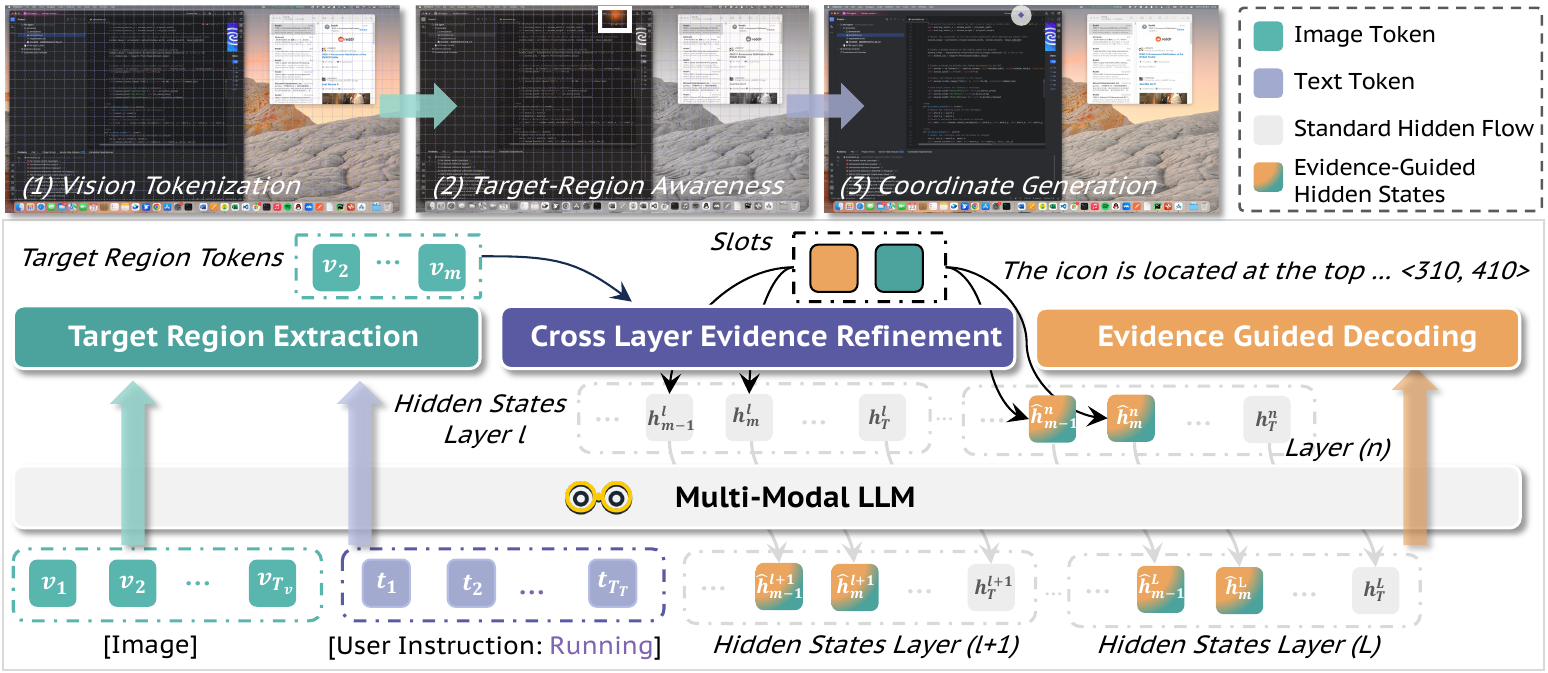}
    \caption{
    \textbf{Overview of \modelname.}
    \modelname bridges intermediate target-region awareness and final coordinate generation within a single forward pass.
    Given a GUI screenshot and an instruction, it identifies a target-aware region from intermediate decoder responses and retrieves fine-grained visual evidence from this region.
    The localized evidence is maintained and refined across decoder layers through a compact evidence workspace, and is then injected into the key/value projections of the hidden states at selected target-region positions.
    }
    \label{fig:method_pipeline}
\end{figure*}

Given a GUI screenshot and a natural-language instruction, the goal of MLLM-based GUI grounding is to generate the target click coordinate autoregressively.
In this work, we introduce \modelname, which turns target-related visual cues emerging within the same forward pass into refined localized evidence for precise point-level coordinate prediction.

The inference pipeline of \modelname is illustrated in Fig.~\ref{fig:method_pipeline}.
We begin by deriving a text-image relevance map from intermediate decoder representations before coordinate decoding, which provides a coarse target-region proposal for retrieving fine-grained vision-encoder features (\S\ref{ssec:roi_extraction}).
Next, a shared Iterative Evidence Adapter refines this localized evidence across decoder layers through a compact evidence workspace (\S\ref{ssec:dual_slot_adapter}).
This workspace maintains evolving evidence states across layers, allowing target-region evidence to be progressively updated and preserved as decoding proceeds.
Finally, the refined evidence is injected into the key/value projections of selected target-region positions, allowing coordinate tokens to access position-sensitive evidence while preserving the original autoregressive decoding interface (\S\ref{ssec:evidence-guided}).

\subsection{Target-Region Evidence Extraction}
\label{ssec:roi_extraction}

Our diagnostic analysis in \S\ref{sec:introduction} shows that intermediate decoder responses can already highlight target-relevant UI regions, even when the final coordinate prediction fails.
We therefore use these responses as self-guided localization cues to determine where fine-grained visual evidence should be extracted for subsequent decoding.

Specifically, given the hidden states $H^{\ell-1}$ before decoder layer $\ell$, we extract the instruction-token states $H_T^{\ell-1}=\{\mathbf{h}^{\ell-1}_{t}\}_{t=1}^{T_T}$ and the decoder-side visual-token states $H_V^{\ell-1}=\{\mathbf{h}^{\ell-1}_{v}\}_{v=1}^{T_V}$.
We first average the instruction-token states into a single text-conditioned query state, and then compute its response to the decoder-side visual tokens using the original query and key projections of layer $\ell$:
\begin{equation}
A_{T \rightarrow V}^{\ell}
=
\mathrm{Softmax}
\left(
\frac{
\bar{Q}_{T}^{\ell}(K_V^{\ell})^\top
}{\sqrt{d_h}}
\right),
\label{eq:text_to_vision_response}
\end{equation}
where $\bar{Q}_{T}^{\ell}$ and $K_V^{\ell}$ are projected from the averaged instruction state and visual-token states, respectively, $d_h$ is the head dimension, and the softmax is applied over visual tokens.
We average $A_{T \rightarrow V}^{\ell}$ across heads and normalize it into a response map.
High-response connected components are converted into bounding boxes as coarse target-region priors for fine-grained evidence extraction.
Details are provided in the Appendix \ref{ssec:target_region}.

Guided by the target-region prior, we retrieve the corresponding final-layer vision-encoder features from the same forward pass.
Unlike the compressed visual tokens passed to the language decoder, these features are extracted before visual token merging and thus preserve richer local layout and appearance details.
We denote them as $X_R \in \mathbb{R}^{N \times U \times d_v}$, where each of the $N$ selected region tokens corresponds to $U$ unmerged visual patches of dimension $d_v$.
In this way, the decoder response specifies where to gather evidence, and the unmerged features provide the local details needed for point-level coordinate prediction.

\subsection{Cross Layer Evidence Refinement}
\label{ssec:dual_slot_adapter}

After extracting regional evidence $X_R$, we then introduce a cross-layer evidence workspace for evidence maintenance and refinement.
By propagating localized visual evidence across decoder layers, the workspace enables progressive refinement and injection of coordinate-relevant cues within a single forward pass.

\noindent\textbf{Cross-layer evidence workspace.}
To enable progressive refinement, we introduce an \emph{Iterative Dual-Slot Evidence Adapter}.
Let $\mathcal{L}=\{\ell_1,\ell_2,\ldots,\ell_K\}$ denote the decoder layers where the adapter is inserted.
The adapter $A_{\theta}$ is shared across these layers, while its evidence state is propagated from one layer to the next.
In this way, the adapter performs recurrent evidence refinement within the original forward pass.
At refinement step $k$, the workspace maintains two learnable evidence slots,
$Z^{(k)}=[\mathbf{z}^{(k)}_1,\mathbf{z}^{(k)}_2]\in\mathbb{R}^{2\times d_s}$.
More details on the slot design are provided in the Appendix \ref{ssec:slot_separation}.

\noindent\textbf{Recurrent target-region evidence aggregation.}
Given the cross-layer workspace, we specify how each slot aggregates regional evidence at decoder layer $\ell_k$.
We flatten $X_R$ into $M=NU$ region tokens, denoted as $\bar{X}_R\in\mathbb{R}^{M\times d_v}$, and project them into keys and values
$\mathbf{K},\mathbf{V}\in\mathbb{R}^{M\times d_s}$.

For each slot $j\in\{1,2\}$, the shared adapter constructs a slot-specific query by combining the current decoding context with the current slot state:
\begin{equation}
\mathbf{q}^{(k)}_j
=
\mathbf{q}^{(k)}_{\mathrm{ctx}}
+
W_z \mathrm{LN}(\mathbf{z}^{(k)}_j)
+
\mathbf{e}_j ,
\label{eq:slot_query}
\end{equation}
where $\mathbf{q}^{(k)}_{\mathrm{ctx}}\in\mathbb{R}^{d_s}$ is derived from the decoder state at layer $\ell_k$ and the instruction representation, $W_z$ is a learnable projection, and $\mathbf{e}_j$ is a learnable slot embedding.
This makes evidence aggregation conditioned on both the current decoding state and the evidence stored in each slot.

To keep aggregation aligned with the target region, we cache a normalized visual log-prior $\log\pi\in\mathbb{R}^{M}$ over the flattened region tokens and reuse it across refinement steps.
The candidate evidence for slot $j$ is obtained by prior-guided attention:
\begin{equation}
\tilde{\mathbf{z}}^{(k)}_j
=
\mathrm{Softmax}
\left(
\frac{\mathbf{q}^{(k)}_j \mathbf{K}^{\top}}{\sqrt{d_s}}
+
\beta_j \log\pi
\right)
\mathbf{V},
\label{eq:prior_guided_aggregation}
\end{equation}
where $\beta_j$ is a learnable slot-specific prior weight.
The query-key term adapts evidence selection to the current slot state, while the visual prior guides aggregation toward the estimated target region.

\noindent\textbf{Gated evidence update.}
After aggregating the candidate evidence $\tilde{\mathbf{z}}^{(k)}_j$, the adapter updates each slot with a learnable gate:
\begin{equation}
\begin{aligned}
\mathbf{g}^{(k)}_j
&=
\sigma\left(W_g[\mathbf{z}^{(k)}_j;\tilde{\mathbf{z}}^{(k)}_j]\right), \\
\mathbf{z}^{(k+1)}_j
&=
\mathbf{z}^{(k)}_j
+
\mathbf{g}^{(k)}_j
\odot
\left(\tilde{\mathbf{z}}^{(k)}_j-\mathbf{z}^{(k)}_j\right).
\end{aligned}
\label{eq:gated_slot_update}
\end{equation}
The gate controls how much candidate evidence is written into the slot, while preserving the evidence from previous layers.
The updated slots are passed to the next inserted layer, enabling cross-layer evidence refinement.

\subsection{Evidence-Guided Coordinate Decoding}
\label{ssec:evidence-guided}
After cross-layer refinement, the evidence slots are injected into decoding through a KV-only strategy.
Specifically, we derive evidence-enhanced states for the target-region positions and use them only to update their key/value projections.
The token sequence and all query projections remain unchanged, so coordinate tokens still attend to the original target-region positions, but receive refined local evidence through their keys and values.

\noindent\textbf{Slot-conditioned evidence fusion.}
At layer $\ell_k$, let
$Z^{(k+1)}=[\mathbf{z}^{(k+1)}_1;\mathbf{z}^{(k+1)}_2]\in\mathbb{R}^{2\times d_s}$
denote the updated evidence workspace, and let $\mathcal{R}_{\mathrm{lm}}$ denote the target-region positions in the language-model sequence.
For each $r\in\mathcal{R}_{\mathrm{lm}}$, we use its decoder state $\mathbf{h}^{(k)}_r$ to attend over the two evidence slots and obtain slot-conditioned evidence $\mathbf{h}^{(k)}_{\mathrm{slot},r}$.
This evidence is fused through a gated residual path:
\begin{equation}
\hat{\mathbf{h}}^{(k)}_r
=
\mathbf{h}^{(k)}_r
+
\eta^{(k)}_r
\odot
W_o \mathbf{h}^{(k)}_{\mathrm{slot},r},
\label{eq:decoder_evidence_fusion}
\end{equation}
where $\eta^{(k)}_r$ is a learned gate controlling the contribution of slot evidence, and $W_o$ maps the slot evidence to the decoder hidden dimension.
The resulting $\hat{\mathbf{h}}^{(k)}_r$ is used only for KV evidence injection.

\noindent\textbf{Target-region KV evidence injection.}
For each target-region position $r\in\mathcal{R}_{\mathrm{lm}}$, we replace its key/value projections with those computed from the evidence-enhanced state:
\begin{equation}
K^{(\ell_k)}_r
\leftarrow
W_K^{(\ell_k)}\hat{\mathbf{h}}^{(k)}_r,
\qquad
V^{(\ell_k)}_r
\leftarrow
W_V^{(\ell_k)}\hat{\mathbf{h}}^{(k)}_r .
\label{eq:kv_only_injection}
\end{equation}
The query projections and the key/value projections of all non-region positions remain unchanged.
This minimally invasive injection preserves the original decoding flow while making the selected target-region positions carry refined, position-specific visual evidence for coordinate prediction.

\begin{table}[!t]
    \centering
    \scriptsize
    \setlength{\tabcolsep}{9.5pt}
    \renewcommand{\arraystretch}{1.0}
    \caption{
    Representative comparison across GUI grounding benchmarks.
    Best and second-best results are highlighted in \besttext{green} and \secondtext{purple}, respectively.
    }
    \label{tab:overall_gui_grounding_comparison}
    \vspace{-0.5em}
    \resizebox{0.98\columnwidth}{!}{
    \begin{tabular}{lccccccc}
        \toprule
        \textbf{Method} 
        & \textbf{Size}
        & \makecell[c]{\textbf{OSW-GR}} 
        & \makecell[c]{\textbf{OSW-G}} 
        & \textbf{SS-V2} 
        & \textbf{SS-Pro} 
        & \textbf{UI-V} 
        & \textbf{MMB-GUI} \\
        \midrule

        Jedi \citep{xie2026scaling}
        & 3B 
        & 61.0 & 50.9 & -- & -- & -- & --\\

        InfiGUI-G1 \citep{infiguig1}
& 3B & -- & -- & -- & 45.2 & 22.0 & -- \\

Ferret-UI Lite \citep{yang2025ferret}
& -- & -- & -- & -- & 53.3 & -- & -- \\
        
        MAI-UI \citep{zhou2025mai}
        & 4B 
        & 69.1 
        & 60.3 
        & 94.8
        & \second{65.1} 
        & \second{37.0} 
        & 85.0 \\

        OS-Atlas \citep{osaltas_and_screenspot_v2}
        & 7B 
        & -- 
        & 27.7 
        & 85.1 
        & -- 
        & 9.0 
        & 41.4 \\

        Aguvis \citep{xu2024aguvis}
        & 7B 
        & -- 
        & 38.7 
        & -- 
        & -- 
        & -- 
        & 45.7 \\

        UGround \citep{uground}
        & 7B 
        & -- 
        & 36.4 
        & 87.7 
        & -- 
        & 12.9 
        & 65.7 \\

        UI-TARS \citep{uitars}
        & 7B 
        & -- 
        & 47.5 
        & 91.6 
        & 35.7 
        & 17.6 
        & -- \\

        UI-TARS-1.5 \citep{ui-tars-15-seed}
        & 7B 
        & 64.2 
        & 52.8 
        & 91.6 
        & 35.7 
        & 22.3 
        & 64.3 \\

        GUI-Actor \citep{wu2026gui}
        & 7B 
        & -- 
        & -- 
        & 92.1 
        & 44.6 
        & -- 
        & 76.5 \\

        SE-GUI \citep{yuan2026se}
        & 7B 
        & -- 
        & -- 
        & 90.8 
        & 47.2 
        & -- 
        & 76.6 \\

        GUI-G$^2$ \citep{tang2026gui}
        & 7B 
        & -- 
        & -- 
        & 93.3 
        & 47.5 
        & -- 
        & 78.8 \\

        UI-Venus \citep{ui_venus}
        & 7B 
        & 61.7 
        & 54.6 
        & 94.1 
        & 50.8 
        & 26.5 
        & 79.9 \\

        GTA1 \citep{GTA1}
        & 7B 
        & 67.7 
        & 60.1 
        & 92.4 
        & 50.1 
        & -- 
        & 78.5 \\

        OpenCUA \citep{opencua}
        & 7B 
        & -- 
        & 53.3 
        & 92.3 
        & 50.0 
        & 29.7 
        & -- \\

        UI-Ins \citep{chen2025ui}
        & 7B 
        & -- 
        & -- 
        & \second{94.9} 
        & 57.0 
        & -- 
        & 83.1 \\

        InfiGUI-G1 \citep{liu2026infigui}
        & 7B 
        & -- 
        & -- 
        & 93.5 
        & 51.9 
        & 26.1 
        & 80.8 \\

        GUI-Owl \citep{ye2025mobile}
        & 7B 
        & -- 
        & 55.9 
        & 92.8 
        & 54.9 
        & -- 
        & 80.5 \\

        Qwen3-VL \citep{Qwen3-VL}
        & 8B 
        & 64.4 
        & 55.7 
        & 92.9 
        & 49.5 
        & 23.3 
        & 81.3 \\

        OpenCUA \citep{wang2026opencua}
        & 32B 
        & \second{70.2} 
        & 59.6 
        & 93.4 
        & 55.3 
        & -- 
        & -- \\

        GUI-Owl \citep{ye2025mobile}
        & 32B 
        & -- 
        & 58.0 
        & 93.2 
        & 58.0 
        & -- 
        & 83.0 \\

        Qwen3-VL \citep{Qwen3-VL}
        & 32B 
        & 69.0 
        & \second{60.6} 
        & 93.0 
        & 54.9 
        & 26.9 
        & \second{85.3} \\

        UI-TARS-DPO \citep{uitars}
        & 72B 
        & -- 
        & 57.1 
        & -- 
        & 38.1 
        & 25.5 
        & 74.3 \\


        \midrule
        \rowcolor{gray!10}
        \textbf{ZoomOnce} 
        & \textbf{4B} 
        & \best{73.1} 
        & \best{64.7} 
        & \best{95.2} 
        & \best{66.2} 
        & \best{40.2} 
        & \best{87.6} \\

        \bottomrule
    \end{tabular}
    }
    \vspace{-0.8em}
\end{table}
 
\subsection{Training Stages}
In the supervised fine-tuning (SFT) stage, we optimize the standard autoregressive cross-entropy loss over the full target sequence, including grounding reasoning and the final coordinate answer.
This adapts the newly introduced evidence pathway while preserving the instruction-following and output-formatting behavior of the base model.
We also apply a lightweight slot separation regularization to reduce redundancy between the two evidence slots, with details provided in the Appendix.

In the reinforcement learning (RL) stage, we initialize from the SFT model and apply GRPO~\citep{shao2024deepseekmath} to further optimize point-level grounding accuracy.
For each instruction, the policy samples multiple responses, and each response is rewarded according to whether its parsed coordinate falls inside the target bounding box.
GRPO estimates relative advantages within each sampled group, encouraging responses with better grounding outcomes beyond token-level imitation.
More training details are provided in the Appendix \ref{ssec:training_obj}.

\section{Experiments}
\label{sec:experiment}

\begin{table}[!t]
    \centering
    \tiny
    \setlength{\tabcolsep}{9.5pt}
    \renewcommand{\arraystretch}{1.0}
    \caption{
    Comparison at matched model scales.
    \besttext{Green} and \secondtext{purple} indicate the best and second-best results within each group.
    \textit{Base SFT+RL} uses the original Qwen3-VL backbone with the same SFT+RL framework and training data, but without our proposed designs.
    $\Delta$ rows show gains/drops over the corresponding baseline.
    The first group compares test-time scaling methods that use zoom/focus operations.  }
    \label{tab:scale_matched_gui_grounding}
    \vspace{-0.5em}
    \resizebox{0.92\columnwidth}{!}{
    \begin{tabular}{lccccc}
        \toprule
        \textbf{Method} 
        & \textbf{Size}
        & \textbf{UI-V}
        & \makecell[c]{\textbf{OSW-GR}} 
        & \makecell[c]{\textbf{OSW-G}} 
        & \textbf{SS-Pro} \\
        \midrule

        \rowcolor{gray!12}
        \multicolumn{6}{l}{\textbf{\textit{Test-time scaling / zoom-based methods}}} \\
        ZoomClick~\citep{jiang2025zoom}
        & 7B 
        & \second{34.0} 
        & -- 
        & -- 
        & \second{65.7} \\
        ZoomUI~\citep{liu2026zoom}
        & 7B 
        & 27.1 
        & -- 
        & 54.2 
        & 52.0 \\
        GUI-Spotlight~\citep{lei2025gui}
        & 7B 
        & 23.4 
        & -- 
        & \second{62.7} 
        & 52.8 \\
        MVP~\citep{zhang2025mvp}
        & 8B 
        & 31.9 
        & \second{72.7} 
        & -- 
        & 65.3 \\
        Region-Focus~\citep{luo2025visual}
        & 72B 
        & -- 
        & -- 
        & -- 
        & 61.6 \\
        \rowcolor{gray!6}
        \textbf{ZoomOnce}
        & \textbf{4B} 
        & \best{40.2}
        & \best{73.1} 
        & \best{64.7} 
        & \best{66.2} \\
        \midrule
        \rowcolor{gray!12}
        \multicolumn{6}{l}{\textbf{\textit{2B models}}} \\
        Qwen3-VL 
        & 2B & 13.7 & 59.6 & 45.7 & 38.3 \\
        \quad + Zoom-In 
        & 2B & 15.8 & 64.1 & 51.2 & 50.9 \\
        Base SFT+RL 
        & 2B & 27.3 & 61.5 & 50.0 & 51.3 \\
        \quad + Zoom-In 
        & 2B 
        & \second{28.9}
        & \second{64.4} 
        & \second{53.2} 
        & \best{56.2} \\
        \rowcolor{gray!6}
        \textbf{\modelname} 
        & \textbf{2B} 
        & \best{30.5}
        & \best{66.8} 
        & \best{53.7} 
        & \second{53.5} \\

        \quad $\Delta$ vs. Base SFT+RL
        & -- 
        & \posdelta{+3.2}
        & \posdelta{+5.3} 
        & \posdelta{+3.7} 
        & \posdelta{+2.2} \\

        \quad $\Delta$ vs. Base SFT+RL + Zoom-In
        & -- 
        & \posdelta{+1.6}
        & \posdelta{+2.4} 
        & \posdelta{+0.5} 
        & \negdelta{-2.7} \\
        \midrule

        \rowcolor{gray!12}
        \multicolumn{6}{l}{\textbf{\textit{4B models}}} \\
        Qwen3-VL 
        & 4B & 24.8 & 65.7 & 57.6 & 53.1 \\
        \quad + Zoom-In 
        & 4B & 26.8 & 71.5 & 61.0 & 63.1 \\
        Base SFT+RL 
        & 4B & 34.0 & 67.2 & 58.3 & 62.6 \\
        \quad + Zoom-In 
        & 4B 
        & \second{36.4}
        & \second{72.5} 
        & \second{62.8} 
        & \best{67.3} \\
        \rowcolor{gray!6}
        \textbf{\modelname} 
        & \textbf{4B} 
        & \best{40.2}
        & \best{73.1} 
        & \best{64.7} 
        & \second{66.2} \\
        \quad $\Delta$ vs. Base SFT+RL
        & -- 
        & \posdelta{+6.2}
        & \posdelta{+5.9} 
        & \posdelta{+6.4} 
        & \posdelta{+3.6} \\

        \quad $\Delta$ vs. Base SFT+RL + Zoom-In
        & -- 
        & \posdelta{+3.8}
        & \posdelta{+0.6} 
        & \posdelta{+1.9} 
        & \negdelta{-1.1} \\
        \bottomrule
    \end{tabular}
    }
\end{table}

\subsection{Experimental Settings}
\noindent\textbf{Training Data.}
Following UI-Ins~\citep{chen2025ui}, we train \modelname on public datasets including OS-Atlas~\citep{osaltas_and_screenspot_v2}, OmniAct~\citep{omniact}, AndroidControl~\citep{android_control}, AMEX~\citep{AMEX}, and AgentNet~\citep{yang2026agentnet}.
We adopt the same data processing pipeline as UI-Ins, resulting in 283K SFT samples and 100K RL samples.
We use Qwen3-VL-Instruct 2B/4B as the backbone model.
More training details are provided in the Appendix \ref{apd:exp_details}.

\noindent\textbf{Metrics and Benchmarks.}
We evaluate \modelname using action accuracy \citep{GTA1}, where a prediction is counted as correct if the predicted coordinate falls inside the target bounding box.
We report results on six GUI grounding benchmarks spanning desktop, mobile, and web interfaces: ScreenSpot-Pro (SS-Pro)~\citep{li2025screenspotpro}, ScreenSpot-V2 (SS-V2)~\citep{osaltas_and_screenspot_v2}, OSWorld-G (OSW-G) and OSWorld-G-Refine (OSW-GR)~\citep{xie2026scaling}, UI-Vision (UI-V)~\citep{ui_vision}, and MMBench-GUI-L2 (MMB-GUI)~\citep{wang2025mmbenchgui}.

\noindent\textbf{Baselines.}
We compare \modelname with representative GUI grounding models, such as MAI-UI~\citep{zhou2025mai}, UI-Ins~\citep{chen2025ui}, GTA1~\citep{GTA1}, UI-TARS~\citep{ui-tars-15-seed}, UI-Venus~\citep{team2026ui}, JEDI~\citep{xie2026scaling}, Qwen3-VL~\citep{Qwen3-VL}, and OS-Atlas~\citep{osaltas_and_screenspot_v2}.

\noindent\textbf{Parameter Settings.}
We attain the target regions from decoder layer 19 and keep the top-3 candidate regions, which cover the ground-truth target in around 55\%--90\% of cases across benchmarks.
We then perform cross-layer evidence refinement at decoder layers 20, 23, 26, and 29 to progressively update localized evidence for coordinate decoding.
More details are provided in the Appendix.

\begin{table}[t!]
\centering
\scriptsize
\setlength{\tabcolsep}{14.5pt}
\renewcommand{\arraystretch}{1.1}
\caption{
Fine-grained comparison on the \textbf{UI-Vision} grounding dataset.
\besttext{Green}/\secondtext{purple} indicates the best/second-best results within each scale group.
}
\label{tab:ui_vision_fine_grained}
\vspace{-1.0em}
\resizebox{0.8\columnwidth}{!}{
\begin{tabular}{lcccc}
\toprule
\textbf{Model} 
& \textbf{Basic} 
& \textbf{Functional} 
& \textbf{Spatial} 
& \textbf{Avg.} \\ 
\midrule

\rowcolor{gray!12}
\multicolumn{5}{l}{\textbf{\textit{Larger models for reference}}} \\
\refcell{Qwen3-VL-32B} \citep{Qwen3-VL}
& \refcell{32.8} & \refcell{34.2} & \refcell{14.7} & \refcell{26.9} \\
\refcell{UI-TARS-72B} \citep{uitars}
& \refcell{31.4} & \refcell{30.5} & \refcell{14.7} & \refcell{25.5} \\
\refcell{UI-Venus-72B} \citep{ui_venus}
& \refcell{45.6} & \refcell{42.3} & \refcell{23.7} & \refcell{36.8} \\ 

\midrule
\rowcolor{gray!12}
\multicolumn{5}{l}{\textbf{\textit{2B/3B models}}} \\
Qwen3-VL-2B \citep{Qwen3-VL}
& 0.0 & 19.2 & 0.1 & 6.2 \\
InfiGUI-G1-3B \citep{infiguig1}
& 31.2 & 28.0 & 8.2 & 22.0 \\
MAI-UI-2B \citep{zhou2025mai}
& \best{41.0} & \best{41.2} & \second{10.4} & \second{30.3} \\
\rowcolor{gray!6}
\textbf{\modelname-2B} (\textcolor{tongyipurple}{Ours})
& \second{37.5} & \second{39.5} & \best{15.3} & \best{30.5} \\

\midrule
\rowcolor{gray!12}
\multicolumn{5}{l}{\textbf{\textit{Recent 4B/7B/8B models}}} \\
Qwen3-VL-8B \citep{Qwen3-VL}
& 25.0 & 27.9 & 1.2 & 17.5 \\
UI-TARS-1.5-7B \citep{ui-tars-15-seed}
& 28.8 & 27.5 & 10.7 & 22.3 \\
InfiGUI-G1-7B \citep{infiguig1}
& 36.2 & 31.9 & 11.5 & 26.1 \\
UI-Venus-7B \citep{ui_venus}
& 36.1 & 32.8 & 11.9 & 26.5 \\
Phi-Ground \citep{phi_ground}
& 36.8 & 37.1 & 7.6 & 27.2 \\
MAI-UI-4B \citep{zhou2025mai}
& \second{47.80} & \second{46.50} & \second{18.40} & \second{37.00} \\
\rowcolor{gray!6}
\textbf{\modelname-4B}  (\textcolor{tongyipurple}{Ours})
& \best{49.15} & \best{47.52} & \best{25.43} & \best{40.24} \\

\bottomrule
\end{tabular}
}
\end{table}

\begin{table*}[t!]
    \centering
    \scriptsize
    \setlength{\tabcolsep}{4.5pt}
    \renewcommand{\arraystretch}{1.1}
    \caption{
    Fine-grained comparison on the \textbf{MMBench-GUI L2} benchmark.
    \besttext{Green}/\secondtext{purple} indicates the best/second-best results within each scale group.
    Larger models are shown in gray for reference.
    }
    \label{tab:mmbench_gui_l2_fine_grained}
    \resizebox{\textwidth}{!}{
    \begin{tabular}{lccccccccccccc}
        \toprule
        \multirow{2}{*}{\textbf{Model}} &
        \multicolumn{2}{c}{\textbf{Windows}} &
        \multicolumn{2}{c}{\textbf{MacOS}} &
        \multicolumn{2}{c}{\textbf{Linux}} &
        \multicolumn{2}{c}{\textbf{iOS}} &
        \multicolumn{2}{c}{\textbf{Android}} &
        \multicolumn{2}{c}{\textbf{Web}} &
        \multirow{2}{*}{\textbf{Avg.}} \\
        \cmidrule(lr){2-3}
        \cmidrule(lr){4-5}
        \cmidrule(lr){6-7}
        \cmidrule(lr){8-9}
        \cmidrule(lr){10-11}
        \cmidrule(lr){12-13}
        & Bas. & Adv. & Bas. & Adv. & Bas. & Adv. & Bas. & Adv. & Bas. & Adv. & Bas. & Adv. & \\
        \midrule

        \rowcolor{gray!12}
        \multicolumn{14}{l}{\textbf{\textit{Larger models for reference}}} \\
        \refcell{GUI-Owl-32B} \citep{ye2025mobile}
            & \refcell{85.6} & \refcell{65.1}
            & \refcell{84.9} & \refcell{67.1}
            & \refcell{77.0} & \refcell{63.3}
            & \refcell{95.2} & \refcell{85.5}
            & \refcell{96.1} & \refcell{87.0}
            & \refcell{95.5} & \refcell{80.8}
            & \refcell{83.0} \\
        \refcell{GTA1-32B} \citep{GTA1}
            & \refcell{82.3} & \refcell{66.9}
            & \refcell{89.0} & \refcell{74.0}
            & \refcell{73.3} & \refcell{52.0}
            & \refcell{96.2} & \refcell{88.2}
            & \refcell{95.8} & \refcell{88.5}
            & \refcell{95.2} & \refcell{79.9}
            & \refcell{83.4} \\
        \refcell{Qwen3-VL-32B} \citep{Qwen3-VL}
            & \refcell{93.4} & \refcell{71.3}
            & \refcell{92.8} & \refcell{74.3}
            & \refcell{78.0} & \refcell{56.1}
            & \refcell{95.5} & \refcell{88.8}
            & \refcell{97.2} & \refcell{88.5}
            & \refcell{92.6} & \refcell{78.6}
            & \refcell{85.3} \\
            
        \midrule
        \rowcolor{gray!12}
        \multicolumn{14}{l}{\textbf{\textit{Recent 4B/7B/8B models}}} \\
        UI-TARS-1.5-7B \citep{ui-tars-15-seed}
            & 68.3 & 39.0
            & 69.0 & 44.5
            & 64.4 & 37.8
            & 88.5 & 69.4
            & 90.5 & 69.3
            & 81.0 & 56.5
            & 64.3 \\
        UGround-V1-7B \citep{uground}
            & 66.8 & 39.0
            & 71.3 & 48.6
            & 56.5 & 31.1
            & 92.7 & 70.9
            & 93.5 & 71.0
            & 88.7 & 64.6
            & 65.7 \\
        GUI-Actor-7B \citep{wu2026gui}
            & 80.8 & 55.1
            & 81.4 & 60.4
            & 64.9 & 41.8
            & 94.3 & 82.7
            & 93.5 & 79.7
            & 89.7 & 72.1
            & 76.5 \\
        SE-GUI-7B \citep{yuan2026se}
            & 77.5 & 57.7
            & 77.1 & 60.7
            & 68.6 & 44.9
            & \second{95.5} & 80.0
            & 95.5 & 83.7
            & 89.7 & 68.8
            & 76.6 \\
        Qwen3-VL-8B \citep{Qwen3-VL}
            & 88.6 & 61.8
            & 85.5 & 69.1
            & 74.9 & 53.1
            & 95.2 & 82.4
            & 95.5 & 84.5
            & \best{96.8} & 72.1
            & 81.3 \\
        GTA1-7B \citep{GTA1}
            & 76.8 & 57.4
            & 80.3 & 63.9
            & 68.6 & 53.6
            & 93.9 & 83.3
            & 96.3 & 84.5
            & 90.3 & 74.7
            & 78.5 \\
        GUI-G$^2$-7B \citep{tang2026gui}
            & 79.7 & 55.1
            & 79.7 & 64.7
            & 69.6 & 50.0
            & 95.2 & 82.7
            & \second{96.6} & 85.4
            & 91.9 & 75.6
            & 78.8 \\
        GUI-Owl-7B \citep{ye2025mobile}
            & 86.4 & 61.8
            & 81.7 & 64.5
            & 74.4 & 61.7
            & 94.9 & 83.0
            & 95.8 & 83.7
            & 93.2 & 72.7
            & 80.5 \\
        InfiGUI-G1-7B \citep{infiguig1}
            & 82.7 & 61.8
            & 83.8 & 63.9
            & 72.3 & 52.0
            & 94.9 & \best{89.4}
            & 95.2 & 85.6
            & 93.5 & 76.3
            & 80.8 \\
        MAI-UI-4B \citep{zhou2025mai}
            & \second{91.9} & \second{72.4}
            & \second{85.2} & \second{74.3}
            & \second{79.1} & \second{63.8}
            & \second{95.5} & 87.3
            & \second{96.6} & \second{87.6}
            & 94.2 & \second{79.6}
            & \second{85.0} \\
        \rowcolor{gray!6}
        \textbf{\modelname-4B} 
            & \best{92.3} & \best{72.8}
            & \best{91.9} & \best{81.2}
            & \best{83.8} & \best{67.9}
            & \best{97.5} & \second{88.2}
            & \best{97.2} & \best{88.7}
            & \second{96.1} & \best{82.1}
            & \best{87.6} \\
        \bottomrule
    \end{tabular}
    }
    \vspace{-0.8em}
\end{table*}

\subsection{Main Results}
\label{ssec:exp_main_results}

\noindent\textbf{Comparison with Recent SOTA Methods.}
Table~\ref{tab:overall_gui_grounding_comparison} reports our main results with the 4B version of \modelname.
With a 4B backbone, \modelname achieves the best accuracy on all six benchmarks, outperforming recent 4B GUI agents and substantially larger 7B/8B/32B/72B models.
Compared with the strongest prior result on each benchmark, \modelname improves OSW-GR, OSW-G, SS-V2, SS-Pro, UI-V, and MMB-GUI by 2.9, 4.1, 0.3, 1.1, 3.2, and 2.3 points, respectively.
We further evaluate a 2B variant to verify the effectiveness of our design, and the results are reported in the following tables and the Appendix.

\noindent\textbf{Scale-Matched Comparison.}
Table~\ref{tab:scale_matched_gui_grounding} provides controlled scale-matched comparisons and compares with recent test-time scaling methods.
In the test-time scaling group, \modelname-4B achieves the best results on all four benchmarks, outperforming zoom/focus-based methods with larger 7B/8B/72B backbones.
This shows that our method improves grounding accuracy without relying on repeated zooming or larger model scale.

Under the same 4B setting, \modelname outperforms the two-pass Base SFT+RL + Zoom-In baseline on UI-V, OSW-GR, and OSW-G by +3.8, +0.6, and +1.9 points, respectively, while being slightly lower on SS-Pro.
This gap is likely caused by the ultra-wide dual-screen samples in SS-Pro, where explicit cropping can better adapt the input to the training distribution.
Nevertheless, \modelname achieves stronger performance on most benchmarks within a single forward pass, showing that its improvements come from effective cross-layer evidence bridging rather than extra supervision or repeated inference.
The 2B results show the same trend, further verifying that our design remains effective under smaller model capacity.

\begin{table}[!t]
    \centering
    \scriptsize
    \setlength{\tabcolsep}{10.5pt}
    \renewcommand{\arraystretch}{1.1}
    \caption{
    Fine-grained comparison on \textbf{OSWorld-G-Refine}. 
    \besttext{Green}/\secondtext{purple} indicates the best/second-best results within each scale group.
    }
    \label{tab:osworld_g_refine_fine_grained}
    \resizebox{0.9\columnwidth}{!}{
    \begin{tabular}{lccccc}
        \toprule
   \textbf{Agent Model} & \makecell[c]{\textbf{Text}\\\textbf{Matching}} & \makecell[c]{\textbf{Element}\\\textbf{Recognition}} & \makecell[c]{\textbf{Layout}\\\textbf{Understanding}} & \makecell[c]{\textbf{Fine-grained}\\\textbf{Manipulation}} & \textbf{Avg} \\
\midrule

        \rowcolor{gray!12}
        \multicolumn{6}{l}{\textbf{\textit{Larger models for reference}}} \\
        \refcell{OpenCUA-32B} \citep{opencua}
        & \refcell{63.2} & \refcell{79.9} & \refcell{84.9} & \refcell{62.1} & \refcell{70.2} \\
        \refcell{Qwen3-VL-32B} \citep{Qwen3-VL}
        & \refcell{77.4} & \refcell{73.6} & \refcell{76.3} & \refcell{57.7} & \refcell{69.0} \\
        \refcell{GTA1-32B} \citep{GTA1}
        & \refcell{63.2} & \refcell{83.6} & \refcell{84.4} & \refcell{70.5} & \refcell{72.2} \\

\midrule
\rowcolor{gray!12}
\multicolumn{6}{l}{\textbf{\textit{2B/3B models}}} \\
Qwen3-VL-2B \citep{Qwen3-VL}
& 69.3 & 60.9 & 69.2 & 45.0 & 57.4 \\
MAI-UI-2B \citep{zhou2025mai}
& \second{70.9} & \second{69.1} & \second{72.7} & \second{47.7} & \second{63.5} \\
\rowcolor{gray!6}
\textbf{\modelname-2B}  (\textcolor{tongyipurple}{Ours})
& \best{73.2} & \best{73.0} & \best{75.5} & \best{52.0} & \best{66.8} \\
        \midrule
        \rowcolor{gray!12}
        \multicolumn{6}{l}{\textbf{\textit{Recent 4B/7B/8B models}}} \\
        UI-TARS-1.5-7B \citep{ui-tars-15-seed}
        & 52.6 & 75.4 & 72.4 & 66.7 & 64.2 \\
        Qwen3-VL-8B \citep{Qwen3-VL}
        & 73.9 & 68.2 & 73.1 & 54.4 & 64.4 \\
        GTA1-7B \citep{GTA1}
        & 63.2 & \second{82.1} & 74.2 & \best{70.5} & 67.7 \\
        MAI-UI-4B \citep{zhou2025mai}
        & \second{84.6} & 80.4 & \second{82.0} & 62.2 & \second{69.1} \\
        \rowcolor{gray!6}
        \textbf{\modelname-4B}  (\textcolor{tongyipurple}{Ours})
        & \best{88.3} & \best{84.6} & \best{85.4} & \second{68.9} & \best{73.1} \\

        \bottomrule
    \end{tabular}
    }
\end{table}

\subsection{Fine-Grained Analysis}
\label{ssec:exp_finegrained_analysis}

To complement the overall results, we further examine category-level breakdowns on representative GUI grounding benchmarks.

\noindent\textbf{General UI grounding.}
UI-Vision evaluates grounding under basic, functional, and spatial instruction types.
As shown in Table~\ref{tab:ui_vision_fine_grained}, \modelname-4B achieves the best results among recent 4B/7B/8B models across all three categories.
The improvement is especially clear on spatial grounding, where \modelname improves over MAI-UI-4B from 18.4 to 25.4.
This indicates that cross-layer evidence bridging helps preserve local target evidence for more precise point-level localization.
The 2B results show a similar trend, where \modelname-2B achieves the best spatial accuracy and the best average result within the 2B/3B group.

\noindent\textbf{Complex desktop grounding.}
OSWorld-G-Refine evaluates desktop grounding scenarios that require text matching, element recognition, layout understanding, and fine-grained manipulation.
As shown in Table~\ref{tab:osworld_g_refine_fine_grained}, \modelname-4B ranks first in text matching, element recognition, and layout understanding, and remains competitive in fine-grained manipulation.
This suggests that \modelname strengthens the visual-semantic alignment and layout-level discrimination required for complex desktop interfaces.
Under the 2B setting, \modelname also achieves the best results across all subcategories.

\noindent\textbf{Cross-platform GUI grounding.}
MMB-GUI covers diverse platforms with both basic and advanced instructions.
As shown in Table~\ref{tab:mmbench_gui_l2_fine_grained}, \modelname-4B achieves the best results in most platform-level subcategories, including both basic and advanced settings on Windows, MacOS, Linux, and Android.
It also performs strongly on Web advanced instructions, where precise localization often requires maintaining fine-grained target evidence until coordinate decoding.
These results demonstrate that \modelname generalizes across operating systems and interface styles, especially in categories that require more than coarse UI recognition.
More fine-grained results on OSWorld-G, SS-Pro, and SS-V2 are provided in the Appendix \ref{ssec:app_add_qresults}.

\begin{figure}[t]
    \centering
    \includegraphics[width=0.7\linewidth]{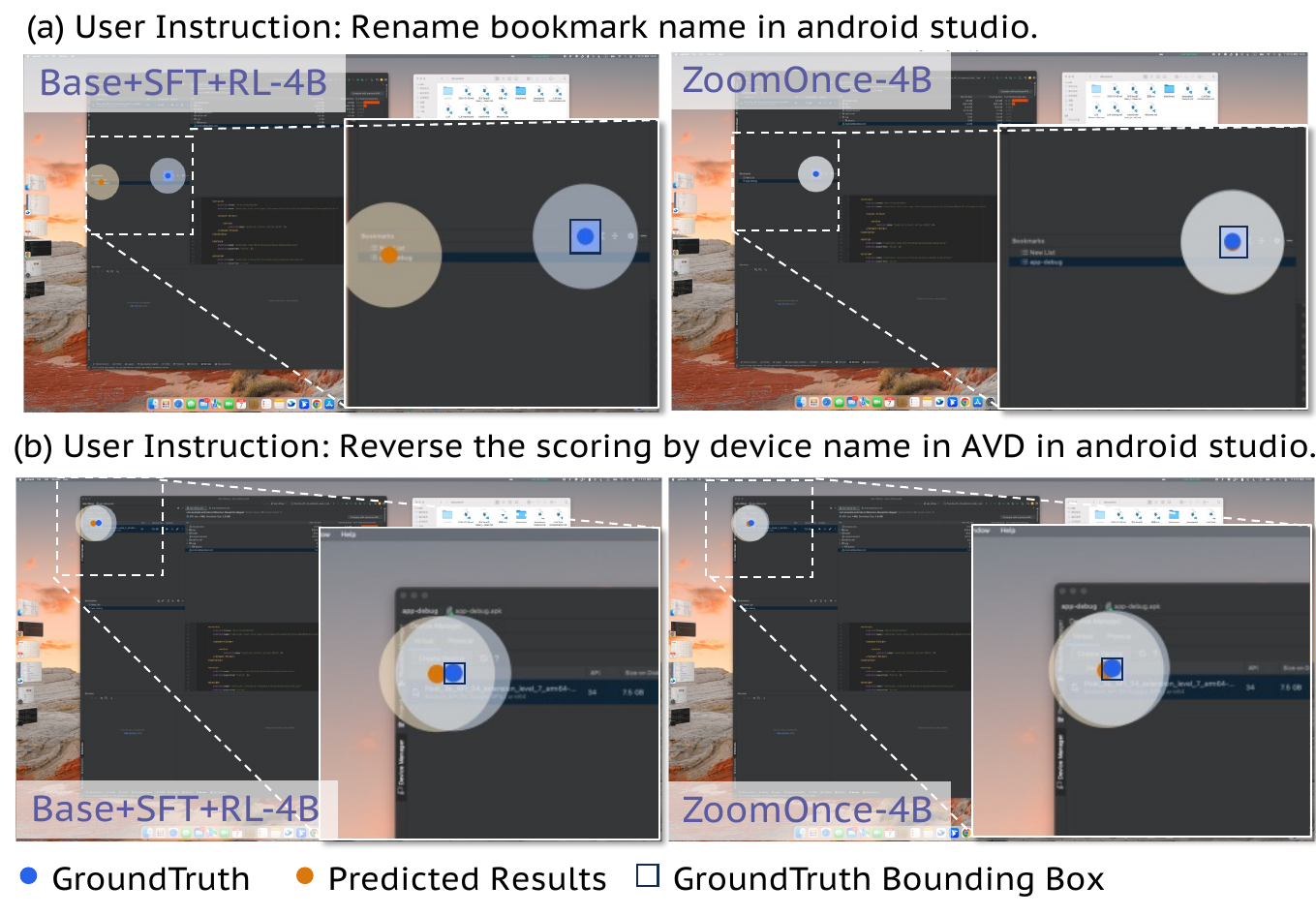}
    \caption{
    Qualitative comparison between \modelname and Base+SFT+RL on ScreenSpot-Pro.
    \modelname achieves more accurate point-level grounding. 
    More examples are provided in the Appendix.
    }
    \vspace{-1.0em}
    \label{fig:visual_case}
\end{figure}

\subsection{Ablation Study}
\label{ssec:ablation_study}

\noindent\textbf{Effect of Cross-Layer Interaction Layers.}
Table~\ref{tab:sft_ablation_study} studies the placement of cross-layer interaction during SFT.
Using two or three interaction layers gives 61.0 and 62.1 weighted accuracy, while our four-layer design at layers 20, 23, 26, and 29 reaches 64.2 with 180.4M trainable parameters.
Densely inserting interaction from layer 20 to the last decoder layer only improves accuracy by 0.1 points, but increases the parameters to 721.8M.
This shows that a compact set of middle-to-late layers is sufficient for effective evidence propagation.

\noindent\textbf{Effect of Slot Number.}
The single-slot and three-slot variants drop to 62.5 and 62.8, respectively, compared with 64.2 from the two-slot design.
This suggests that one slot lacks sufficient capacity to separate target and contextual evidence, while extra slots may cause redundant aggregation.
The two-slot design therefore offers a compact and effective target-context evidence workspace.

\subsection{Accuracy--Efficiency Trade-off}
\label{ssec:accuracy_efficiency}

\begin{figure}[t]
    \centering
    \includegraphics[width=0.7\linewidth]{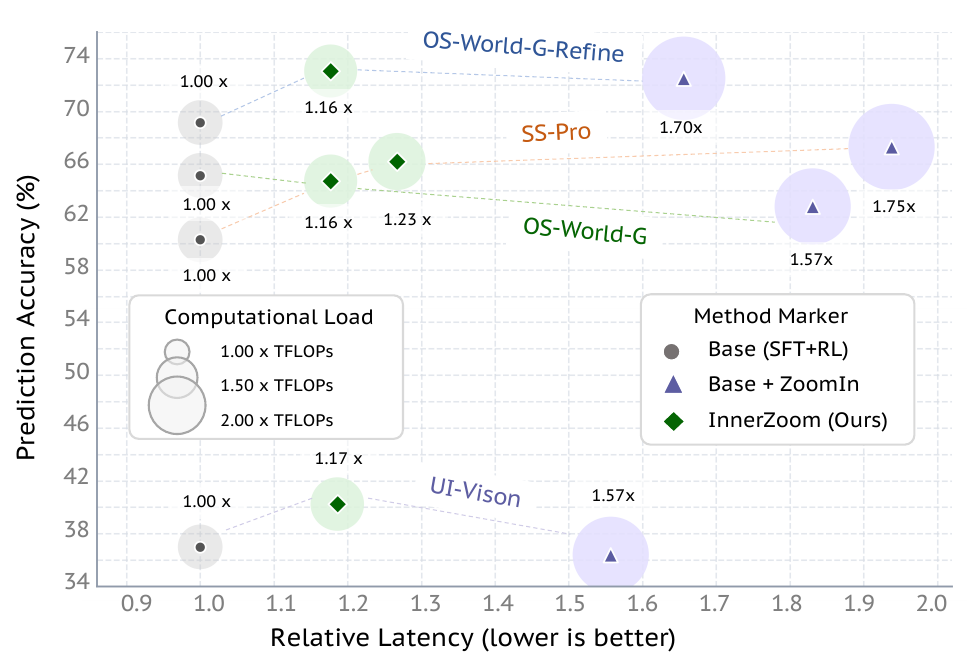}
    \caption{
    Accuracy--efficiency trade-off under the scale-matched 4B setting.
    The x-axis denotes relative end-to-end latency normalized by Base SFT+RL on the same benchmark, and the y-axis reports action accuracy.
    Bubble size represents relative TFLOPs, also normalized by the corresponding Base SFT+RL result.
    Compared with two-pass Zoom-In, \modelname achieves comparable accuracy with substantially lower latency and computational cost on most benchmarks.
    }
    \label{fig:efficiency}
\end{figure}

We further analyze the accuracy--efficiency trade-off under the 4B scale-matched setting.
Fig.~\ref{fig:efficiency} compares Base SFT+RL, Base SFT+RL + Zoom-In, and \modelname on OSW-GR, OSW-G, SS-Pro, and UI-V.
Since these benchmarks differ in image resolution, we normalize latency and TFLOPs within each benchmark using Base SFT+RL as the reference.
Detailed measurement and estimation protocols are provided in the Appendix \ref{ssec:app_latency_tflpos}.

As shown in Fig.~\ref{fig:efficiency}, Base SFT+RL + Zoom-In relies on a conditional second forward pass, leading to substantially higher overhead with $1.56$--$1.94\times$ latency and $1.57$--$1.75\times$ TFLOPs.
In contrast, \modelname achieves stronger accuracy on most benchmarks while staying close to the base model, requiring only $1.18$--$1.27\times$ latency and $1.16$--$1.23\times$ TFLOPs.
Compared with Zoom-In, this reduces latency by $23.8$--$35.7\%$ and TFLOPs by $26.0$--$32.0\%$, cutting most of the additional overhead introduced by repeated inference.
With this much smaller computational cost, \modelname still outperforms Zoom-In on OSW-GR, OSW-G, and UI-V by +0.6, +1.9, and +3.8 points, respectively.
These results show that reusing localized evidence within a single forward pass provides a more favorable accuracy--efficiency trade-off than external zoom-based reprocessing.

\begin{table}[!t]
    \centering
    \scriptsize
    \setlength{\tabcolsep}{11.8pt}
    \renewcommand{\arraystretch}{1.1}
    \caption{
    SFT-stage ablation study on cross-layer interaction layers and slot number.
    Weighted Avg. Acc. is sample-count weighted prediction accuracy over six datasets.
    $\Delta$ is computed relative to our final design.
}
    \label{tab:sft_ablation_study}
    \resizebox{0.8\columnwidth}{!}{
    \begin{tabular}{llcccc}
        \toprule
        \textbf{Ablation}
        & \makecell[c]{\textbf{Cross-layer}\\\textbf{Interaction}}
        & \makecell[c]{\textbf{\#}\\\textbf{Slots}}
        & \makecell[c]{\textbf{Trainable}\\\textbf{Params} $\downarrow$}
        & \makecell[c]{\textbf{Weighted}\\\textbf{Avg. Acc.} $\uparrow$}
        & $\Delta$ \\
        \midrule

        \rowcolor{gray!12}
        \multicolumn{6}{l}{\textbf{\textit{Effect of cross-layer interaction layers}}} \\
        \ding{172}
        & 20, 23
        & 2
        & \best{90.2M}
        & 61.0
        & $-3.2$ \\
        \ding{173}
        & 20, 23, 26
        & 2
        & \second{135.3M}
        & 62.1
        & $-2.1$ \\
        \rowcolor{gray!6}
        \ding{173}
        & \textbf{20, 23, 26, 29}
        & \textbf{2}
        & \textbf{180.4M}
        & \second{\textbf{64.2}}
        & -- \\
        \ding{174}
        & 20-last layer
        & 2
        & 721.8M
        & \best{64.3}
        & $+0.1$ \\

        \midrule
        \rowcolor{gray!12}
        \multicolumn{6}{l}{\textbf{\textit{Effect of slot number}}} \\
        Single-slot
        & 20, 23, 26, 29
        & 1
        & --
        & 62.5
        & $-1.7$ \\
        \rowcolor{gray!6}
        \textbf{Two-slot}
        & \textbf{20, 23, 26, 29}
        & \textbf{2}
        & \textbf{180.4M}
        & \best{\textbf{64.2}}
        & -- \\
        Three-slot
        & 20, 23, 26, 29
        & 3
        & --
        & \second{62.8}
        & $-1.4$ \\

        \bottomrule
    \end{tabular}
    }
\end{table}

\subsection{Qualitative and Failure Analysis}
\label{ssec:qualitative_failure_analysis}

\noindent\textbf{Qualitative Results.} Fig.~\ref{fig:visual_case} presents a qualitative comparison between Base SFT+RL and \modelname on SS-Pro.
\modelname produces more accurate point-level predictions in dense UI regions with visually similar neighboring elements, showing that localized evidence bridging helps refine coordinate prediction beyond coarse target-region awareness (see more in App. \ref{ssec:app_add_qresults}.).

\noindent\textbf{Failure cases.}
The remaining errors mainly arise from semantically challenging instructions, ambiguous text-to-image grounding, and interference from on-screen text.
These cases suggest that although \modelname improves fine-grained localization, robust domain-specific instruction understanding and command-interface text disentanglement remain open challenges (see more in App. \ref{ssec:qualitative_failure_analysis}.).

\section{Related Work}
\label{sec:related_work}

\subsection{GUI Grounding for GUI Agents.}
\label{sec:ssec_relatedworkGUIGrounding}

GUI grounding is a core capability of GUI agents, enabling them to map natural-language instructions to executable interface actions, such as clicking, typing, and navigation.
Early works \citep{hsieh2026zonui, zhao2026points, wu2024gui, wang2026opencua, gu2025ui, xue2026evocua}, such as SeeClick \citep{cheng2024seeclick}, trained visual GUI agents to locate instruction-relevant UI elements directly from screenshots.
Later methods improve GUI grounding from both data and agent-modeling perspectives.
UGround \citep{qian2025uground} scales GUI visual grounding with large-scale synthetic data, OS-Atlas \citep{wu2025atlas} builds a large cross-platform GUI grounding corpus, and ShowUI \citep{lin2025showui} introduces UI-aware visual token selection for vision-language-action modeling.
Recent GUI agents and grounding models, including UI-TARS \citep{qin2025ui, wang2025ui}, MAI-UI \citep{zhou2025mai}, GUI-Owl \citep{xu2026mobile}, and Aria-UI \citep{yang2025aria}, further integrate grounding into end-to-end, native, or pure-vision GUI agents.

Recent works explore GUI grounding from different optimization perspectives.
UI-Ins \citep{chen2025ui} improves instruction-level reasoning, RULER \citep{wang2025improving} improves coordinate prediction by modeling position-to-coordinate mapping, and GUI-Actor \citep{wu2026gui} replaces direct coordinate generation with action-region prediction.
Meanwhile, SE-GUI \citep{yuan2026se}, GUI-G2 \citep{tang2026gui}, GUI-RCPO \citep{du2026test}, InfiGUI-G1 \citep{liu2026infigui}, and GuirlVG \citep{kang2025guirlvg} improve training or inference through dense spatial rewards, region consistency, RL exploration, or stabilized reinforcement fine-tuning.
Despite these advances, these methods largely overlook the inherent mismatch between the generative MLLM paradigm and the fine-grained spatial alignment required for point-level GUI grounding.
We addresse this gap by transforming target-region cues emerging within a single forward pass into localized visual evidence for final coordinate decoding.


Zoom-based methods improve fine-grained GUI grounding by revisiting local regions at higher effective resolution.
RegionFocus \citep{luo2025visual} formalizes dynamic zoom-in as visual test-time scaling, while ZoomClick \citep{jiang2025zoom} studies zooming as a training-free prior for GUI grounding.
DiMo-GUI \citep{wu2025dimo} further combines dynamic visual grounding with modality-aware reasoning over textual and iconic UI elements.
Recent Zoom-based frameworks make this process more adaptive.
UI-Zoomer \citep{tang2026ui} triggers zoom-in according to prediction uncertainty, AdaZoom-GUI \citep{pei2026adazoom} combines instruction refinement with conditional zoom-in, and ZoomUI progressively anchors instructions to interface elements through latent thinking and attention-guided zooming.

Beyond zooming, MVP \citep{zhang2025mvp} mitigates coordinate prediction instability by aggregating predictions from multiple attention-guided views.
Chain-of-Ground \citep{li2025chain} and Iterative Narrowing \citep{nguyen2024improved} refine grounding through multi-step visual reasoning or progressive cropping, while GUI-Spotlight \citep{lei2025gui} and GUI-Eyes \citep{chen2026gui} introduce tool-augmented focus refinement and active visual perception.
Despite different designs, these methods \citep{lin2026happens, lee2025reguide, zhou2025gui, chen2026v2p, zhangmanicog, liu2026zoom, liu2026zoom} commonly aim to supplement global GUI understanding with localized, high-resolution visual evidence for fine-grained coordinate prediction.
However, they typically acquire such evidence through additional test-time computation.
In contrast, our work reuses localized target-region evidence within a single forward pass to support final point-level coordinate decoding.

\section{Conclusion}
\label{sec:conclusion}

We presented \modelname, a single-forward cross-layer evidence bridging framework for precise GUI grounding.
Our analysis reveals an evidence-to-coordinate bottleneck, where intermediate decoder layers already form useful target-region evidence but fail to reliably convert it into final click coordinates.
\modelname addresses this gap by preserving and refining intermediate target evidence across decoder layers and making it available for coordinate decoding.
Experiments suggest that this evidence-preserving design improves GUI grounding accuracy while maintaining efficient single-forward inference, providing an effective alternative to two-pass zoom-in refinement.

\section{Limitations}
\label{sec:limitations}

Despite its strong performance, \modelname still has several limitations.
First, its evidence extraction depends on target-region cues emerging from intermediate decoder layers.
When these internal responses are incomplete or biased, the selected region may limit subsequent evidence refinement.
Since region selection is only indirectly optimized during training, future work could explore learnable or weakly supervised mechanisms for more robust evidence extraction.

Second, although \modelname achieves strong results on high-resolution benchmarks such as ScreenSpot-Pro, explicit visual re-observation may still be complementary in extreme cases.
For ultra-wide or dual-screen interfaces, a two-pass zoom-in strategy can provide higher effective resolution over the target area, offering visual details that are difficult to fully recover from the original forward pass.
This suggests that single-forward evidence bridging and zoom-based re-observation address different aspects of precise grounding, and future work may combine them adaptively for more challenging high-resolution GUI scenarios.

\clearpage

\appendix

\section{Appendix}
\label{sec:appendix}
\startcontents[appendix]
\printcontents[appendix]{}{1}{\setcounter{tocdepth}{3}}

\subsection{Method Details}
\label{sec:app_methods}
This section provides additional implementation details of our method. 
In particular, we elaborate on the details of locating text and visual tokens in the input sequence, constructing a heatmap from the text-to-vision response, selecting ROIs through connected-component analysis, and mapping the expanded regions back to token indices.

\subsubsection{Target-Region Proposal and Feature Retrieval}
\label{ssec:target_region}

\noindent\textbf{Text-to-Vision Heatmap Construction.}
To obtain a target-region proposal from the model's own intermediate responses, we first locate the user-instruction tokens and visual tokens in the input sequence.
Specifically, we identify the user instruction span using the special tokens $<\vert\texttt{im\_start}\vert>$ and $<\vert\texttt{im\_end}\vert>$, and locate the visual-token span using $<\vert\texttt{vision\_start}\vert>$ and $<\vert\texttt{vision\_end}\vert>$. 
Since different prompt templates or preprocessing pipelines may place the visual tokens either before or after the user instruction, and batched inputs may be left- or right-padded depending on the training or inference setting, we employ the attention mask provided with the model inputs to identify the first valid token position. 
This mask distinguishes real input tokens from padding tokens, allowing us to shift the parsed relative spans to their absolute positions in the full input sequence.
The parsed relative positions are then shifted to absolute indices in the full input sequence, ensuring that subsequent attention hooks access the correct text and visual-token ranges.

Our target region proposal only requires a text-to-vision heatmap from one selected intermediate decoder layer. 
Current decoder backbones often support eager attention \citep{wolf2019huggingface}, PyTorch SDPA, and FlashAttention-style kernels \citep{dao2022flashattention}, which differ in whether intermediate attention maps are exposed. 
Under the standard eager-attention output interface, enabling attention outputs typically materializes full sequence-to-sequence attention matrices across decoder layers, even though we only need the text-to-vision response from one layer. 
This causes unnecessary memory overhead and may slow down training by disabling more efficient fused attention kernels. 
In contrast, PyTorch SDPA returns only the attention output, while FlashAttention-style kernels avoid materializing the full attention matrix to reduce memory traffic.
Thus, directly extracting the required intermediate attention evidence from the backbone is either memory-intensive or incompatible during training.

To preserve the benefits of optimized attention backends while still obtaining the required attention evidence, we recompute the attention response only during target-region proposal generation. 
For the selected intermediate decoder layer, we reuse its Q/K projections to compute a restricted attention response from the user-text tokens to the visual tokens. 
We first average the projected queries of all user-text tokens to obtain an instruction-level query for each attention head. 
To reduce memory usage, we split the visual-key sequence into 256-token chunks along the key dimension, compute the logits between the instruction-level query and each chunk of visual keys, concatenate the chunk-wise logits, and apply softmax over the visual-token dimension. 
The resulting response assigns each visual token a target-relevance score with respect to the user instruction. 
These scores are aggregated across attention heads, normalized, and reshaped into a two-dimensional visual grid to form the text-to-vision heatmap. 
The heatmap captures the target-region awareness that has already emerged before coordinate generation and serves as the basis for subsequent connected-component selection.

\noindent\textbf{Connected Component Selection.}
Given the two-dimensional text-to-vision heatmap, we obtain coarse target-region proposals via connected-component analysis. 
Specifically, we first threshold the heatmap with a fixed quantile threshold \(q_{\mathrm{thr}}=0.90\), retaining the top 10\% most responsive positions as foreground pixels. 
We then apply the 8-neighbor connected-component algorithm to group adjacent foreground pixels into candidate regions.

For each connected component, we compute its area, total heat response, mean response, and minimum enclosing bounding box. 
Then the components are ranked by a score that combines response strength and region size:
\[
s_c = \sum_{p \in c} h_p \cdot |c|^{\alpha},
\]
where \(h_p\) is the heatmap value at position \(p\), \(|c|\) is the component area, and \(\alpha=0.7\). This score favors regions that are both highly responsive and spatially coherent. The highest-scoring component, or the top-\(k\) components when multiple regions are used, is selected as the target-region proposal.

\begin{table*}[t]
\centering
\caption{SFT training hyperparameters.}
\vspace{-0.5em}
\label{tab:sft_hyperparams}
\begingroup
\small
\setlength{\tabcolsep}{5.5pt}          
\renewcommand{\arraystretch}{1.1}   
\resizebox{0.95\textwidth}{!}{%
\small
\begin{tabular}{lccc}
\toprule
\textbf{Hyperparameter} 
& \textbf{Stage 1} 
& \textbf{Stage 2} 
& \textbf{Stage 3} \\
\midrule
Purpose 
& Evidence warm-up 
& Joint training 
& Decoder adaptation \\
Duration 
& 0.032 epoch
& 0.160 epoch 
& 0.808 epoch \\
Learning rate 
& $5{\times}10^{-6}$ 
& $3{\times}10^{-6}\!\rightarrow\!3{\times}10^{-7}$ 
& $2{\times}10^{-5}\!\rightarrow\!2{\times}10^{-6}$ \\
LR schedule 
& Cosine 
& Cosine 
& Cosine \\
Warmup ratio 
& 0.10 
& 0.05 
& 0.05 \\
Weight decay 
& 0.0 
& 0.01 
& 0.01 \\
Trainable modules 
& Adapter only 
& Adapter + decoder layers 
& Decoder + MLP + adapter \\
Max grad norm 
& 1.0 
& 1.0 
& 1.0 \\
Effective batch size
& 256 
& 256 
& 256 \\
Compute resources 
& 256 NVIDIA H20 GPUs
& 256 NVIDIA H20 GPUs
& 256 NVIDIA H20 GPUs \\
\bottomrule
\end{tabular}%
}
\endgroup
\end{table*}

\noindent\textbf{Bounding Box Expansion and Token Index Mapping.}
After selecting a connected component, we compute its minimum enclosing rectangle and use it as the base bounding box for the target region.
Since the attention heatmap often covers only the most responsive part of a target widget, such as the icon center, a text fragment, or a small region inside a button, directly using this box may miss useful boundary and layout cues. We therefore expand the box with a fixed ratio to include the target boundary and nearby context. Specifically, given a box with width \(w\) and height \(h\), we expand it by \(\lfloor \max(w,h) \cdot r_{\mathrm{pad}} \rfloor\) patch positions on each side, where \(r_{\mathrm{pad}}=0.30\) in our implementation. 
The expanded box is clipped to the valid grid to avoid out-of-bound indices.

This expansion is used only as a fixed feature-retrieval heuristic, not as an additional learned module. The same expansion ratio is used across all experiments rather than being tuned for individual datasets or benchmarks. 
After expansion, we map the box back to token indices. For each position \((r,c)\) in the two-dimensional visual grid, its flattened visual-token index is \(r \times W + c\). 
By adding the start position \(v_{\mathrm{start}}\) of the visual-token span in the input sequence, we obtain the corresponding target-region positions in the language-model sequence. 
The resulting token set is used both to retrieve fine-grained pre-merge visual features from the vision encoder and to identify the positions where KV-only evidence injection is applied.

\noindent\textbf{Top-\(k\) Region Union.}
In some cases, target-related responses may be split into multiple local components, for example, when a widget contains both an icon and a text label, or when high responses appear around different parts of the target boundary. 
To improve coverage, we optionally keep the top-\(k\) connected components according to the component score. By default, we use \(k=1\). For each selected component, we compute and expand its bounding box as described above. We then collect all token indices inside the expanded boxes, remove duplicates, sort them, and discard indices outside the visual-token span \([v_{\mathrm{start}}, v_{\mathrm{end}}]\).

The final unioned token set serves as the target-region token set. It connects the proposal stage with the following evidence pathway in two ways: it specifies where to retrieve fine-grained visual features from the vision encoder, and it determines which language-model positions receive KV-only evidence injection during coordinate decoding.

\subsubsection{Slot Separation Regularization}
\label{ssec:slot_separation}

The dual-slot workspace is designed to maintain two complementary evidence states, namely a \emph{focus slot} and a \emph{context slot}. 
However, similar initial query directions may cause the two slots to attend to similar visual patches and gradually converge to redundant representations during training. 
This slot collapse degenerates the dual-slot workspace into two nearly identical evidence vectors, weakening its ability to separate click-relevant local cues from contextual information. 
To mitigate this issue, we encourage slot diversity from both initialization and optimization.

\noindent\textbf{Orthogonal Initialization.}
We first break the symmetry between the two slots through orthogonal initialization. 
The learnable slot query embeddings are initialized with approximately orthogonal directions, allowing the focus and context slots to start from distinct query biases and induce different attention patterns from the first forward pass. 
The initial cross-layer slot states are also initialized as small orthogonal vectors rather than zeros. 
This avoids identical early gated updates while keeping the newly introduced evidence pathway nearly transparent at the beginning of training.

\begin{table}[t]
\centering
\caption{RL training hyperparameters.}
\label{tab:rl_hyperparams}

\small
\setlength{\tabcolsep}{2pt}
\renewcommand{\arraystretch}{1.0}

\begin{tabularx}{0.92\columnwidth}{@{}>{\raggedright\arraybackslash}X>{\raggedright\arraybackslash}X@{}}
\toprule
\textbf{Hyperparameter} & \textbf{Value} \\
\midrule
Initialization & SFT checkpoint \\
Rollouts per prompt & 8 \\
Rollout engine & HuggingFace Transformers-based generation \\
Sampling temperature & 1.0 \\
Top-$p$ & 1.0 \\
Actor micro-batch size & 2 per GPU \\
Log-prob micro-batch size & 2 per GPU \\
Actor strategy & FSDP2 \\
Adapter LR multiplier & 1$\times$ \\
Total epochs & 1 \\
Batch size & 512 \\
Compute resources & 256 NVIDIA H20 GPUs \\
\bottomrule
\end{tabularx}

\vspace{-0.8em}
\end{table}

\subsubsection{Training Objective}
\label{ssec:training_obj}
We train the model in two stages. 
The first stage uses supervised fine-tuning to learn coordinate generation and stabilize the dual-slot evidence pathway. 
The second stage applies GRPO to directly optimize point-level grounding accuracy.

\noindent\textbf{Supervised Fine-Tuning.}
During supervised fine-tuning, we optimize the model with the standard autoregressive cross-entropy loss and an auxiliary slot-separation regularizer. 
For each adapter insertion layer \(\ell \in \mathcal{L}_{\mathrm{inj}}\), we apply the stop-gradient inner-product separation to the updated \emph{focus} and \emph{context} slots, denoted as \(\mathcal{L}_{\mathrm{z\text{-}sep}}^{(\ell)}\). 
The separation loss is averaged over all insertion layers:
\begin{equation}
\mathcal{L}_{\mathrm{sep}}
=
\frac{1}{|\mathcal{L}_{\mathrm{inj}}|}
\sum_{\ell\in\mathcal{L}_{\mathrm{inj}}}
\mathcal{L}_{\mathrm{z\text{-}sep}}^{(\ell)} .
\label{eq:sep-loss}
\end{equation}
Overall, the SFT objective is formulated as:
\begin{equation}
\mathcal{L}_{\mathrm{SFT}}
=
\mathcal{L}_{\mathrm{CE}}
+
\lambda_{\mathrm{sep}}\mathcal{L}_{\mathrm{sep}},
\label{eq:sft-loss}
\end{equation}
where \(\mathcal{L}_{\mathrm{CE}}\) is the autoregressive cross-entropy loss over the target sequence, and \(\lambda_{\mathrm{sep}}=0.02\) controls the strength of the slot-separation regularization.

\noindent\textbf{GRPO Optimization.}
After supervised fine-tuning, we further optimize the model with GRPO to directly improve point-level grounding accuracy as \citep{GTA1,zhou2025mai}. 
For each input \(x\), the policy samples a group of \(G\) responses \(\{y_i\}_{i=1}^{G}\). 
Each response is parsed into a predicted coordinate \((u_i,v_i)\), and receives a binary grounding reward:
\begin{equation}
R_i =
\mathbb{I}\left[(u_i,v_i)\in\mathcal{B}\right],
\qquad
\hat{A}_i =
\frac{R_i-\bar{R}}{\sigma_R+\epsilon_{\mathrm{adv}}},
\label{eq:grpo-reward}
\end{equation}
where \(\mathcal{B}\) denotes the ground-truth target bounding box, \(\hat{A}_i\) is the group-normalized advantage, \(\sigma_R\) is the standard deviation of rewards within the group, and \(\epsilon\) is a small constant for numerical stability. Responses that cannot be parsed into valid coordinates are assigned zero reward.

Following GRPO \citep{shao2024deepseekmath}, we estimate group-wise advantages and optimize the policy with a clipped PPO-style objective \citep{schulman2017proximal}.
\begin{equation}
\begin{aligned}
\mathcal{L}_{\mathrm{GRPO}}
&=
-\frac{1}{G}
\sum_{i=1}^{G}
\frac{1}{T_i}
\sum_{t=1}^{T_i}
\min \Big(
\rho_{i,t}\hat{A}_i, \\
&
\mathrm{clip}
\left(
\rho_{i,t},
1-\epsilon_{\mathrm{clip}},
1+\epsilon_{\mathrm{clip}}
\right)
\hat{A}_i
\Big).
\end{aligned}
\label{eq:grpo-loss}
\end{equation}
where \(T_i\) is the response length, 
\(\rho_{i,t}=\frac{\pi_{\theta}(y_{i,t}\mid x,y_{i,<t})}{\pi_{\mathrm{old}}(y_{i,t}\mid x,y_{i,<t})}\) 
is the token-level policy ratio, and \(\epsilon_{\mathrm{clip}}\) is the clipping threshold. 
This objective encourages responses whose predicted coordinates are more accurate than other samples in the same group, thereby aligning optimization with the final grounding outcome.

\begin{table}[!t]
    \centering
    \tiny
    \setlength{\tabcolsep}{11.0pt}
    \renewcommand{\arraystretch}{1.0}
    \caption{
    Fine-grained comparison on \textbf{OSWorld-G}. 
    \besttext{Green}/\secondtext{purple} indicates the best/second-best results within each scale group.
    }
    \vspace{-0.8em}
    \label{tab:osworld_g_fine_grained}
    \resizebox{\columnwidth}{!}{
    \begin{tabular}{lccccc}
        \toprule
   \textbf{Agent Model} & \makecell[c]{\textbf{Text}\\\textbf{Matching}} & \makecell[c]{\textbf{Element}\\\textbf{Recognition}} & \makecell[c]{\textbf{Layout}\\\textbf{Understanding}} & \makecell[c]{\textbf{Fine-grained}\\\textbf{Manipulation}} & \textbf{Avg} \\
\midrule
\rowcolor{gray!12}
\multicolumn{6}{l}{\textbf{\textit{2B/3B models}}} \\
Qwen3-VL-2B 
& 61.7 & 45.8 & \second{54.2} & 39.6 & 45.9 \\
Jedi-3B
& \best{67.4} & 53.0 & 53.8 & \best{44.3} & 50.9 \\
MAI-UI-2B 
& \second{62.8} & \second{56.7} & \best{59.3} & \second{40.3} & \second{52.0} \\
\rowcolor{gray!6}
\textbf{\modelname-2B}
& 62.1 & \best{59.7} & \best{59.3} & 39.5 & \best{53.7} \\

\midrule
\rowcolor{gray!12}
\multicolumn{6}{l}{\textbf{\textit{Recent 4B/7B/8B models}}} \\
Qwen3-VL-8B 
& 69.0 & 55.5 & 59.7 & 47.7 & 54.8 \\
GTA1-7B 
& 42.1 & 65.7 & 62.7 & 56.1 & 55.1 \\
GUI-Owl-7B 
& 64.8 & 63.6 & 61.3 & 41.0 & 55.9 \\
UI-Venus-7B  
& 74.6 & 60.5 & 61.5 & 45.5 & 58.8 \\
MAI-UI-4B 
& \second{78.8} & \second{68.6} & \second{68.6} & \second{57.8} & \second{60.3} \\
\rowcolor{gray!6}
\textbf{\modelname-4B}
& \best{84.2} & \best{75.5} & \best{74.1} & \best{64.4} & \best{64.7} \\

        \bottomrule
    \end{tabular}
    }
    \vspace{-1.0em}
\end{table}

\subsection{Implementation Details of Diagnostic Attention Intervention}
\label{app:attention_intervention}
We use attention intervention only for the diagnostic analysis in Fig.~\ref{fig:teaser}.
The goal is to test whether the high-response regions identified from intermediate decoder layers contain decision-relevant evidence for final coordinate prediction.
Given the original attention logits, we add a token-wise bias $B$ before softmax.
A positive bias increases the post-softmax attention weight of the corresponding tokens, while a negative bias suppresses it.
Therefore, assigning positive bias to the selected region and negative bias to other visual tokens allows us to selectively amplify target-related evidence.
We obtain the intervention region using the same ROI extraction procedure as in Sec.~\ref{ssec:roi_extraction}.

For the main intervention, we adopt a hard bias strategy.
Tokens inside the selected target region receive $\beta_{\mathrm{in}}>0$, while tokens outside the region receive $\beta_{\mathrm{out}}<0$.
In our diagnostic experiments, we set $\beta_{\mathrm{in}}=5.5$ and $\beta_{\mathrm{out}}=-5.5$.
To rule out the effect of generic attention perturbation, we further introduce a random attention intervention.
Specifically, we randomly select the same number of visual tokens as the ROI tokens and treat them as a pseudo-target region, assigning them the same positive bias $\beta_{\mathrm{in}}$ while assigning $\beta_{\mathrm{out}}$ to the remaining tokens.
Thus, the target-region-based intervention and the random intervention use the same amplification budget.
The performance gap between them indicates whether the selected intermediate regions provide decision-relevant evidence rather than merely benefiting from arbitrary attention modulation.

\subsection{Experiment Details}
\label{apd:exp_details}

\noindent\textbf{Data and Preprocessing.}
We follow the data construction, image preprocessing, coordinate normalization, and evaluation protocol of UI-Ins \citep{chen2025ui} and MAI-UI \citep{zhou2025mai}. 
All samples are converted into a unified GUI grounding format, where the model takes a GUI screenshot and a natural-language instruction as input and generates the target click coordinate. 

\noindent\textbf{Output Format.}
The model is trained to generate a complete grounding response, including grounding reasoning and the final coordinate answer. 
The final answer follows the JSON format \texttt{\{"coordinate": [x, y]\}}, where the coordinate is normalized to the \([0,1000]\) range along the image width and height.

\noindent\textbf{SFT Training Framework.}
SFT is implemented with HuggingFace Transformers and Accelerate \citep{wolf2019huggingface}, together with DeepSpeed ZeRO-2 \citep{rajbhandari2020zero} for distributed training. 
We use bfloat16 precision and FlashAttention2 \citep{dao2022flashattention} to reduce memory cost under high-resolution GUI inputs, and enable gradient checkpointing for long-context training. 
During SFT, the proposed evidence pathway remains active, including the target-region proposal, iterative dual-slot evidence refinement, and KV-only evidence injection, allowing the model to learn the coordinate prediction with the same mechanism at inference time.

\noindent\textbf{Three-Stage SFT Schedule.}
To stabilize the newly introduced evidence pathway, we use a three-stage supervised fine-tuning schedule.
First, we warm up the evidence pathway by mainly training the evidence adapter conditioned on the target-region proposal. 
This allows the model to adapt to target-region feature retrieval, dual-slot refinement, and KV-only evidence injection, while limiting perturbation to the backbone. 
Second, we jointly train the evidence adapter and selected decoder layers. 
This enables the decoder to better incorporate refined target-region evidence when generating grounding responses. 
In the third stage, we further adapt the decoder to improve its coordination prediction accuracy with the evidence pathway. 

\noindent\textbf{RL Training Framework.}
RL training is implemented with Verl \citep{sheng2024hybridflow}, Ray \citep{moritz2018ray}, and FSDP2 \citep{zhao2023pytorch}. 
For each prompt, the policy samples multiple rollouts and computes group-relative advantages from the binary click reward. 
We update the actor with the clipped GRPO objective, with the same evidence-injection pathway enabled during both rollout generation and policy updating. 
This stage directly optimizes point-level grounding under the target-box inclusion criterion, complementing the token-level imitation objective of SFT.

\noindent\textbf{Hyperparameter Summary.}
Detailed hyperparameter settings for SFT and RL are summarized in Table \ref{tab:sft_hyperparams} and Table \ref{tab:rl_hyperparams}, respectively.

\begin{table*}[!t]
    \centering
    \scriptsize
    \setlength{\tabcolsep}{5.5pt}
    \renewcommand{\arraystretch}{1.3}
    \caption{
    Fine-grained comparison on \textbf{ScreenSpot-Pro}. 
    \besttext{Green}/\secondtext{purple} indicates the best/second-best results within each scale group.
    Larger models are shown in gray for reference.
    }
    \vspace{-0.6em}
    \label{tab:sspro_fine_grained}
    \resizebox{\textwidth}{!}{
    \begin{tabular}{lccccccccccccc}
        \toprule
        \multirow{2}{*}{\textbf{Model}} &
        \multicolumn{2}{c}{\textbf{CAD}} &
        \multicolumn{2}{c}{\textbf{Dev.}} &
        \multicolumn{2}{c}{\textbf{Creative}} &
        \multicolumn{2}{c}{\textbf{Scientific}} &
        \multicolumn{2}{c}{\textbf{Office}} &
        \multicolumn{2}{c}{\textbf{OS}} &
        \multirow{2}{*}{\textbf{Avg.}} \\
        \cmidrule(lr){2-3}
        \cmidrule(lr){4-5}
        \cmidrule(lr){6-7}
        \cmidrule(lr){8-9}
        \cmidrule(lr){10-11}
        \cmidrule(lr){12-13}
        & Text & Icon & Text & Icon & Text & Icon & Text & Icon & Text & Icon & Text & Icon & \\
        \midrule

        \rowcolor{gray!15}
        \multicolumn{14}{l}{\textbf{\textit{Larger models for reference}}} \\
        \refcell{Qwen3-VL-32B~\citep{Qwen3-VL}}
        & \refcell{60.4} & \refcell{28.1} 
        & \refcell{69.5} & \refcell{22.1} 
        & \refcell{75.8} & \refcell{25.2} 
        & \refcell{84.7} & \refcell{25.5}
        & \refcell{85.9} & \refcell{43.4} 
        & \refcell{62.6} & \refcell{15.7} 
        & \refcell{54.9} \\

        \refcell{GUI-Owl-32B~\citep{ye2025mobile}}
        & \refcell{62.4} & \refcell{28.1} 
        & \refcell{84.4} & \refcell{39.3} 
        & \refcell{65.2} & \refcell{18.2} 
        & \refcell{82.6} & \refcell{39.1}
        & \refcell{81.4} & \refcell{39.6} 
        & \refcell{70.1} & \refcell{36.0} 
        & \refcell{58.0} \\

        \refcell{GTA1-32B~\citep{GTA1}}
        & \refcell{43.7} & \refcell{23.4} 
        & \refcell{82.5} & \refcell{28.3} 
        & \refcell{69.2} & \refcell{14.7} 
        & \refcell{79.9} & \refcell{31.8}
        & \refcell{80.8} & \refcell{43.4} 
        & \refcell{70.1} & \refcell{32.6} 
        & \refcell{63.6} \\

        \refcell{UI-Venus-72B~\citep{ui_venus}}
        & \refcell{66.5} & \refcell{29.7} 
        & \refcell{84.4} & \refcell{33.1} 
        & \refcell{73.2} & \refcell{30.8} 
        & \refcell{84.7} & \refcell{42.7}
        & \refcell{83.1} & \refcell{60.4} 
        & \refcell{75.7} & \refcell{36.0} 
        & \refcell{61.9} \\




        

        
\rowcolor{gray!15}
\multicolumn{14}{l}{\textbf{\textit{Recent 4B/7B/8B models}}} \\
Qwen3-VL-8B~\citep{Qwen3-VL}
& 46.7 & 10.9 & 79.2 & 23.4 & 68.2 & 14.0 & 73.6 & 30.0
& 76.3 & 30.2 & 65.4 & 21.3 & 49.9 \\

GTA1-7B~\citep{GTA1}
& 53.3 & 17.2 & 66.9 & 20.7 & 62.6 & 18.9 & 76.4 & 31.8
& 82.5 & \second{50.9} & 48.6 & 25.9 & 50.1 \\

UI-Venus-7B~\citep{ui_venus}
& 60.4 & 21.9 & 74.7 & 24.1 & 63.1 & 14.7 & 76.4 & 31.8
& 75.7 & 41.5 & 49.5 & 22.5 & 50.8 \\

InfiGUI-G1-7B~\citep{infiguig1}
& 57.4 & 23.4 & 74.7 & 24.1 & 64.6 & 18.2 
& \second{80.6} & 31.8
& 75.7 & 39.6 & 57.0 & 29.2 & 51.9 \\

GUI-Owl-7B~\citep{ye2025mobile}
& 64.5 & 21.9 
& 76.6 & 31.0 
& 59.6 & 27.3 
& 79.1 & 37.3
& 77.4 & 39.6 
& 59.8 & 33.7 
& 54.9 \\

MAI-UI-4B~\citep{zhou2025mai}
& \second{69.0} & \second{31.3}
& \second{80.5} & \best{53.8}
& \second{71.7} & \second{34.3}
& \best{85.4} & \second{39.1}
& \second{90.4} & 49.1
& \second{78.5} & \second{50.6}
& \second{65.1} \\


\rowcolor{gray!6}
\textbf{\modelname-4B}
& \best{\textbf{69.2}} & \best{\textbf{32.5}} 
& \best{\textbf{82.2}} & \second{\textbf{44.6}}
& \best{\textbf{73.2}} & \best{\textbf{36.7}} 
& 79.7 & \best{\textbf{47.3}} 
& \best{\textbf{91.3}} & \best{\textbf{57.1}} 
& \best{\textbf{79.7}} & \best{\textbf{57.7}} 
& \best{\textbf{66.2}} \\

        \bottomrule
    \end{tabular}
    }
    \vspace{-0.8em}
\end{table*}

\begin{table}[!t]
    \centering
    \scriptsize
    \setlength{\tabcolsep}{17.5pt}
    \renewcommand{\arraystretch}{1.0}
    \caption{
    Fine-grained comparison on \textbf{ScreenSpot-V2}. 
    \besttext{Green}/\secondtext{purple} indicates the best/second-best results within each scale group.
    Larger models are shown in gray for reference.
    }
    \label{tab:screenspot_v2_fine_grained}
    \resizebox{\columnwidth}{!}{
    \begin{tabular}{lccccccc}
        \toprule
        \multirow{2}{*}{\textbf{Model}} 
        & \multicolumn{2}{c}{\textbf{Mobile}} 
        & \multicolumn{2}{c}{\textbf{Desktop}} 
        & \multicolumn{2}{c}{\textbf{Web}} 
        & \multirow{2}{*}{\textbf{Avg.}}  \\
        \cmidrule(lr){2-3} 
        \cmidrule(lr){4-5} 
        \cmidrule(lr){6-7}
        & Text & Icon & Text & Icon & Text & Icon &  \\
        \midrule

\rowcolor{gray!12}
\multicolumn{8}{l}{\textbf{\textit{Larger models for reference}}} \\
\refcell{GUI-Owl-32B}
& \refcell{98.6} & \refcell{90.0} & \refcell{97.9} & \refcell{87.8}
& \refcell{94.4} & \refcell{86.7} & \refcell{93.2} \\
\refcell{GTA1-32B}
& \refcell{99.7} & \refcell{90.5} & \refcell{99.0} & \refcell{94.3}
& \refcell{95.7} & \refcell{90.1} & \refcell{95.2} \\
\refcell{UI-Venus-72B}
& \refcell{99.7} & \refcell{93.8} & \refcell{95.9} & \refcell{90.0}
& \refcell{96.2} & \refcell{92.6} & \refcell{95.3} \\

\midrule
\rowcolor{gray!12}
\multicolumn{8}{l}{\textbf{\textit{2B/3B models}}} \\
Qwen3-VL-2B 
& 95.5 & 82.0 & 95.4 & 73.6 & 89.7 & 76.4 & 86.7 \\
Jedi-3B
& \second{96.6} & 81.5 & 96.9 & 78.6 & 88.5 & 83.7 & 88.6 \\
MAI-UI-2B 
& \best{99.3} & \best{87.2} & \second{97.4} & \second{88.6} & \second{94.0} & \second{84.7} & \second{92.5} \\
\rowcolor{gray!6}
\textbf{\modelname-2B}
& \best{99.3} & \second{84.1} & \best{98.5} & \best{90.3} & \best{95.9} & \best{85.6} & \best{93.0} \\

\midrule
\rowcolor{gray!12}
\multicolumn{8}{l}{\textbf{\textit{Recent 4B/7B/8B models}}} \\
Qwen3-VL-8B 
& \second{97.9} & 84.8 & 95.9 & 87.9 & 95.7 & 83.7 & 91.7 \\
GTA1-7B 
& \best{99.0} & 88.6 & 94.9 & 89.3 & 92.3 & 86.7 & 92.4 \\
GUI-Owl-7B 
& \best{99.0} & \best{92.4} & 96.9 & 85.0 & 93.6 & 85.2 & 92.8 \\
UI-Venus-7B 
& \best{99.0} & \second{90.0} & 96.9 & 90.7 & 96.2 & 88.7 & 94.1 \\
MAI-UI-4B 
& \best{99.0} & 89.6 & \second{98.5} & \second{92.1} & \second{96.6} & 90.6 & \second{94.8} \\
\rowcolor{gray!6}
\textbf{\modelname-4B}
& \best{99.0} & 87.7 & \best{99.5} & \best{95.0} & \best{97.4} & \best{91.1} & \best{95.2} \\

        \bottomrule
    \end{tabular}
    }
    \vspace{-0.8em}
\end{table}

\subsection{More Experimental Results}
\label{sec:app_more_exp}

\subsubsection{Latency Measurement and TFLOPs Estimation.}
\label{ssec:app_latency_tflpos}
For the accuracy--efficiency analysis, we compare Base SFT+RL, Base SFT+RL + Zoom-In, and \modelname under the same 4B scale-matched setting.
Since different benchmarks have different image resolutions, pixel budgets, and input/output token lengths, we report latency and TFLOPs as relative values normalized by the corresponding Base SFT+RL result on the same benchmark.

\noindent\textbf{Latency measurement.}
Latency is measured as end-to-end milliseconds per sample, including image preprocessing, model generation, coordinate decoding, and output parsing.
All timing results are obtained on a single NVIDIA H20 GPU with 96GB memory, using CUDA 12.3, PyTorch 2.3.0a0+ebedce2, and bfloat16 precision.
We use the same decoding configuration as the main evaluation, with \texttt{max\_new\_tokens=128}, greedy decoding, and \texttt{do\_sample=False}.
Generation is performed sample by sample rather than with tensor-batched inference to ensure consistent per-sample timing.

For Base SFT+RL, latency corresponds to one forward generation pass.
For Base SFT+RL + Zoom-In, latency includes the second zoom-in pass only when it is triggered, so the measured overhead reflects the actual trigger rate rather than a fixed $2\times$ cost.
For \modelname, latency includes all additional operations introduced by our method, including visual evidence extraction and cross-layer evidence adaptation.
Relative latency is computed by dividing each method's measured latency by the Base SFT+RL latency on the same benchmark.

\noindent\textbf{TFLOPs estimation.}
TFLOPs are analytically estimated rather than measured with hardware profilers.
We estimate the computational cost from the Qwen3-VL-4B architecture and the average vision pixels, input tokens, and output tokens of each benchmark.
For Base SFT+RL, the estimate includes the vision encoder, language-model prefill, and autoregressive decoding.
For Base SFT+RL + Zoom-In, we scale the single-pass cost by the expected number of forward passes, \emph{i.e.}, one initial pass plus the measured zoom-trigger rate.
For \modelname, the estimate includes the base forward computation and the additional costs of target region evidence extraction and the inserted cross-layer evidence adapters.
Relative TFLOPs are computed by normalizing each method's estimated TFLOPs by the Base SFT+RL TFLOPs on the same benchmark.
This per-benchmark normalization accounts for dataset-specific differences in resolution, pixel budget, and sequence length.

\subsubsection{More Fine-grained Results.}
\label{ssec:app_more_fine_results}

Tables~\ref{tab:osworld_g_fine_grained}, \ref{tab:sspro_fine_grained}, and~\ref{tab:screenspot_v2_fine_grained} provide additional fine-grained comparisons on OSWorld-G, ScreenSpot-Pro, and ScreenSpot-V2.
These results further verify the effectiveness of \modelname across desktop interaction, high-resolution professional interfaces, and general GUI grounding scenarios.

On OSWorld-G, \modelname-4B achieves the best result across all four subcategories, including text matching, element recognition, layout understanding, and fine-grained manipulation.
Compared with MAI-UI-4B, \modelname improves the average score from 60.3 to 64.7, with clear gains in element recognition and layout understanding.
This suggests that cross-layer evidence bridging helps preserve target-relevant visual evidence for both semantic recognition and layout-sensitive grounding.
Under the 2B setting, \modelname also achieves the best average score, further showing that the proposed design remains effective under limited model capacity.

On ScreenSpot-Pro, \modelname-4B obtains the best average score of 66.2, surpassing MAI-UI-4B and larger reference models.
It achieves the best result in 10 out of 12 text/icon subcategories and the best or second-best result in 11 out of 12 subcategories across CAD, development, creative, scientific, office, and OS.
The improvements are especially clear on dense icon-based categories such as scientific and office, where precise grounding requires distinguishing visually similar neighboring elements.
These results show that \modelname remains robust on high-resolution professional GUIs with dense layouts and small targets.

On ScreenSpot-V2, \modelname-4B achieves the best average result within the recent 4B/7B/8B group and remains comparable to much larger reference models.
It performs particularly well on desktop and web splits, achieving the best results on Desktop Text, Desktop Icon, Web Text, and Web Icon.
The 2B variant also achieves the best average score within the 2B/3B group, with strong performance on desktop and web categories.
Overall, these results demonstrate that \modelname consistently improves GUI grounding across task types, resolutions, and interface styles by converting intermediate target-region evidence into more accurate point-level coordinate predictions.
\subsubsection{Additional Qualitative Results.}
\label{ssec:app_add_qresults}

\begin{figure*}
    \centering
    \includegraphics[width=0.88\linewidth]{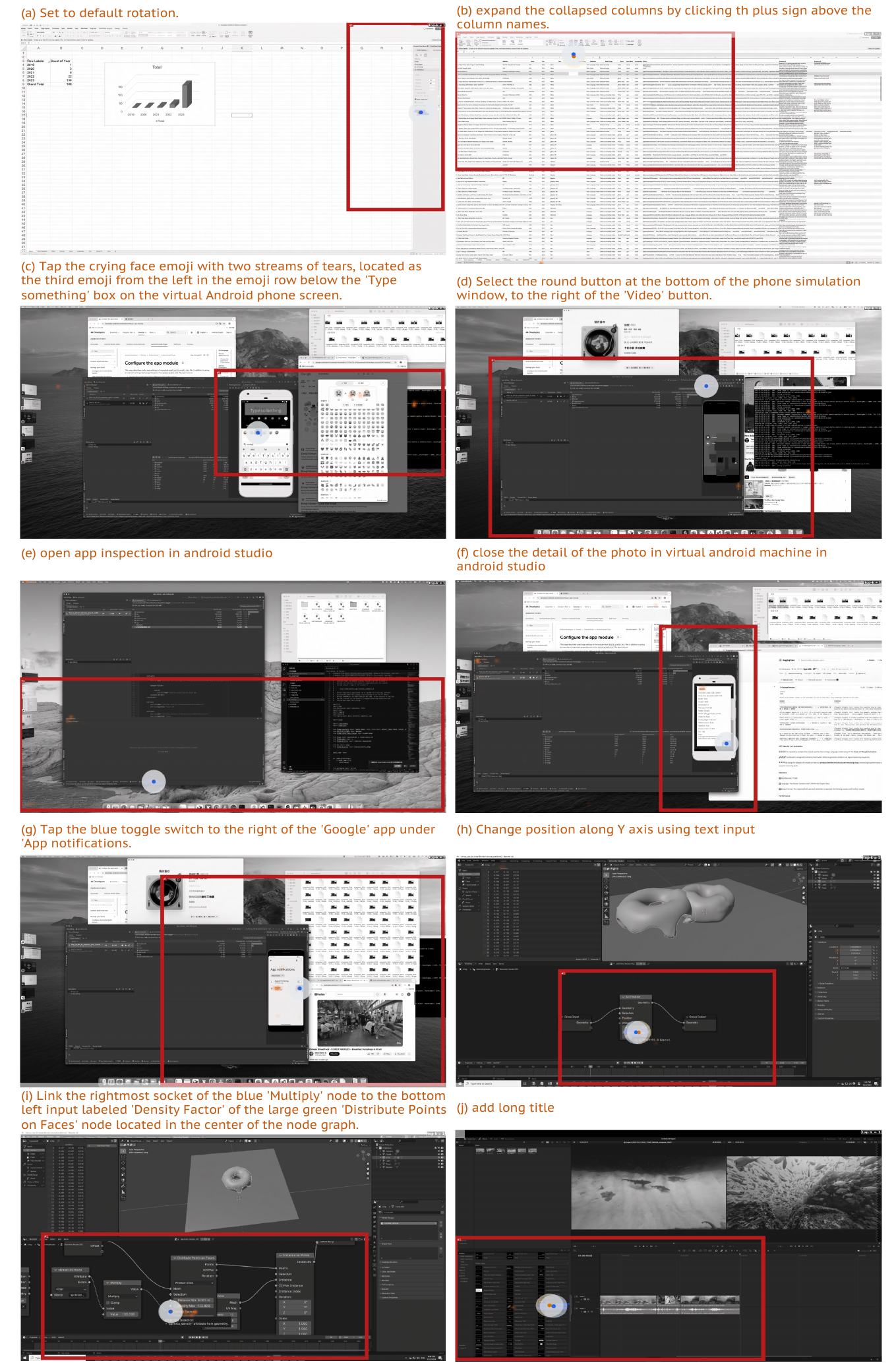}
    \caption{More visualizations on ScreenSpot-Pro. 
    For better readability, we convert the visualization images to grayscale. The red boxes indicate the target regions identified by our method, the orange points denote the predicted results, and the blue points denote the ground-truth points.
    }
    \label{fig:visual_sspro}
\end{figure*}

\begin{figure*}
    \centering
    \includegraphics[width=0.9\linewidth]{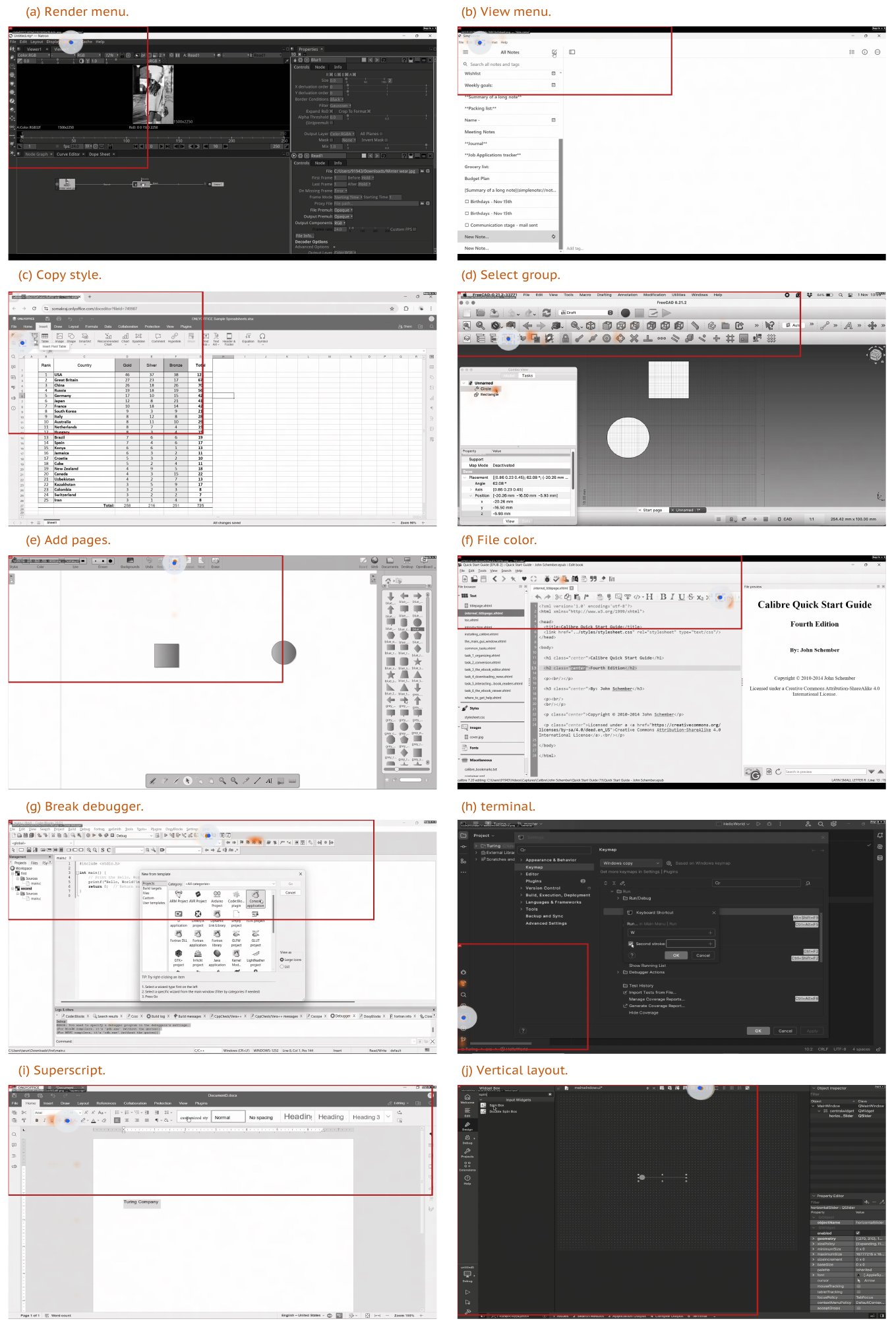}
    \vspace{-0.5em}
    \caption{More visualizations on UI-Vision. 
    For better readability, we convert the visualization images to grayscale. The red boxes indicate the target regions identified by our method, the orange points denote the predicted results, and the blue points denote the ground-truth points.
    }
    \label{fig:visual_ui-vision}
\end{figure*}

Figs.~\ref{fig:visual_sspro} and~\ref{fig:visual_ui-vision} provide additional qualitative examples on SS-Pro and UI-V.
The SS-Pro examples cover high-resolution professional interfaces with dense layouts, small icons, tool panels, and domain-specific controls.
These cases are challenging because the target is often surrounded by visually similar neighboring elements, and the instruction may refer to a fine-grained tool, button, menu item, or functional region.
As shown in Fig.~\ref{fig:visual_sspro}, \modelname produces accurate point-level predictions in these crowded regions, suggesting that cross-layer evidence bridging helps preserve and reuse local visual cues.
We also visualize the dilated union of the estimated target regions with red boxes.
These regions remain spatially compact while preserving sufficient local context around the predicted target, providing effective visual evidence for subsequent point-level grounding.

The UI-V examples further demonstrate the generality of \modelname across broader UI grounding scenarios.
The targets include menu entries, text-formatting controls, debugging buttons, file-related operations, terminal regions, and layout-related elements.
As shown in Fig.~\ref{fig:visual_ui-vision}, \modelname generalizes to diverse interface layouts, indicating that the proposed evidence-bridging method is not limited to a specific UI type.

\subsubsection{Failure analysis.}
\begin{figure*}
    \centering
    \includegraphics[width=1.0\linewidth]{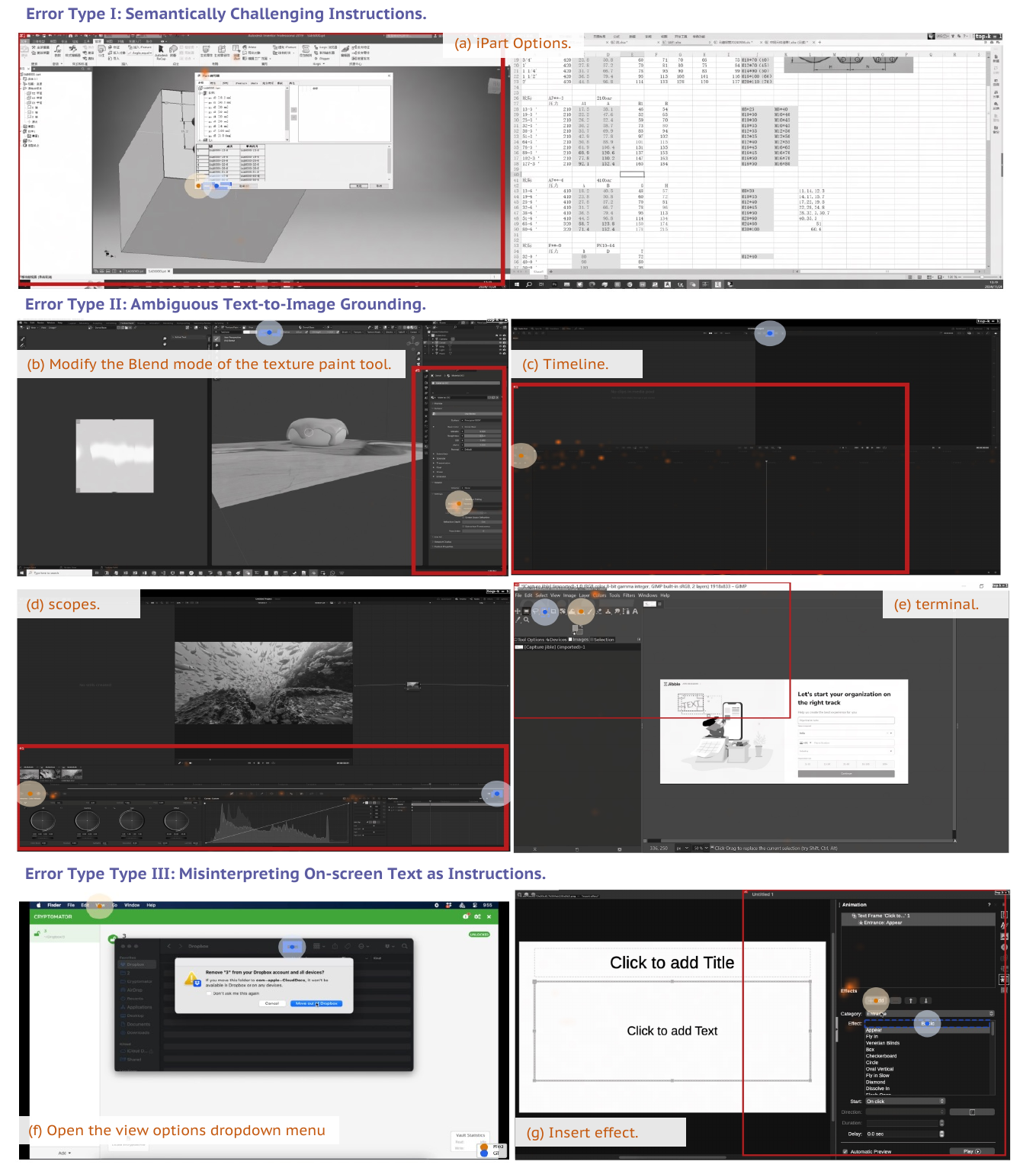}
    \caption{
    Failure case analysis. 
    We identify three representative error types: 
    \textbf{Type I: Semantically Challenging Instructions}, where abstract or indirect instructions make the intended target difficult to infer; 
    \textbf{Type II: Ambiguous Text-to-Image Grounding}, where the instruction is understandable but cannot be clearly mapped to a unique visual element; 
    and \textbf{Type III: Misinterpreting On-screen Text as Instructions}, where textual content in the screenshot distracts the model and is incorrectly treated as part of the user instruction. 
    For better readability, screenshots are shown in grayscale. Red boxes indicate the target regions identified by our method, orange points denote predicted click locations, and blue points denote ground-truth points.
    }
    \label{fig:failure_case}
\end{figure*}

Fig.~\ref{fig:failure_case} summarizes representative failure cases.
We observe three main types of remaining errors.

\noindent\textbf{Type I: Semantically challenging instructions.}
Some failures arise from domain-specific or abstract instructions, such as locating software-specific options or modifying specialized tool settings.
In these cases, the main difficulty is not only point-level localization, but also understanding the operational meaning of the instruction within a particular application.
When the model lacks sufficient domain knowledge about the tool hierarchy or function semantics, the extracted local evidence may still be associated with an incorrect region.

\noindent\textbf{Type II: Ambiguous text-to-image grounding.}
Another common failure occurs when the instruction is understandable, but the corresponding visual target is ambiguous.
This often happens in dense professional interfaces, where multiple regions may partially match the instruction, or the target is small, low-contrast, or visually similar to nearby elements.
Such ambiguity makes it difficult to determine which local evidence should be emphasized during coordinate decoding.

\noindent\textbf{Type III: Misinterpreting on-screen text as user instructions.}
The model may also be distracted by salient on-screen text, especially when the interface contains instruction-like labels, placeholders, or menu items.
In such cases, the model may incorrectly treat visible UI text as part of the user command, leading to attention on task-irrelevant regions.
This suggests that future work may benefit from stronger disentanglement between user instructions and interface text, as well as more explicit modeling of actionable UI elements.


\clearpage
{\small
\bibliographystyle{plainnat}
\bibliography{Bib}
}

\end{document}